\makeatletter
\@ifundefined{isChecklistMainFile}{
  \newif\ifreproStandalone
  \reproStandalonetrue
}{
  \newif\ifreproStandalone
  \reproStandalonefalse
}
\makeatother

\documentclass[letterpaper]{article} % DO NOT CHANGE THIS
\usepackage{aaai2026}  % DO NOT CHANGE THIS 
\usepackage{xcolor}

\usepackage{times}  % DO NOT CHANGE THIS
\usepackage{helvet}  % DO NOT CHANGE THIS
\usepackage{courier}  % DO NOT CHANGE THIS
\usepackage[hyphens]{url}  % DO NOT CHANGE THIS
\usepackage{graphicx} % DO NOT CHANGE THIS
\usepackage{natbib}  % DO NOT CHANGE THIS AND DO NOT ADD ANY OPTIONS TO IT
\usepackage{caption} % DO NOT CHANGE THIS AND DO NOT ADD ANY OPTIONS TO IT
\usepackage{multirow}
\usepackage{threeparttable}
\usepackage{subfig}
\usepackage{amsmath}
\usepackage{amssymb}
\usepackage{booktabs}
\usepackage{algorithm}
\usepackage{algorithmic}

\usepackage{newfloat}
\usepackage{listings}
\DeclareCaptionStyle{ruled}{labelfont=normalfont,labelsep=colon,strut=off} % DO NOT CHANGE THIS
\floatstyle{ruled}
\newfloat{listing}{tb}{lst}{}
\floatname{listing}{Listing}
\title{EMIT: Enhancing MLLMs for Industrial Anomaly Detection via Difficulty-Aware GRPO}
\author{
    Wei Guan\textsuperscript{\rm 1,2}, 
    Jun Lan\textsuperscript{\rm 1}, 
    Jian Cao\textsuperscript{\rm 2}\thanks{Corresponding author.},
    Hao Tan\textsuperscript{\rm 1,3},
    Huijia Zhu\textsuperscript{\rm 1},
    Weiqiang Wang\textsuperscript{\rm 1}
}
\affiliations{
     \textsuperscript{\rm 1} Ant Group \\
    \textsuperscript{\rm 2} Department of Computer Science and Engineering, Shanghai Jiao Tong University \\
     \textsuperscript{\rm 3} Institute of Automation, Chinese Academy of Sciences
    \{guan-wei, cao-jian\}@sjtu.edu.cn, \{yelan.lj, huijia.zhj\}@antgroup.com, tanhao2023@ia.ac.cn,  wang.weiqiang@gmail.com
}

\usepackage{bibentry}
\begin{document}

\maketitle

\begin{abstract}
Industrial anomaly detection (IAD) plays a crucial role in maintaining the safety and reliability of manufacturing systems. While multimodal large language models (MLLMs) show strong vision-language reasoning abilities, their effectiveness in IAD remains limited without domain-specific adaptation. In this work, we propose EMIT, a unified framework that enhances MLLMs for IAD via difficulty-aware group relative policy optimization (GRPO). 
EMIT constructs a multi-task IAD dataset and utilizes GPT-generated object text descriptions to compensate for missing defective images. 
For few-shot anomaly detection, it integrates a soft prompt and heatmap-guided contrastive embeddings derived from patch-level comparisons. 
To better handle difficult data samples, i.e., cases where the MLLM struggles to generate correct answers, we propose a difficulty-aware GRPO that extends the original GRPO by incorporating a response resampling strategy to ensure the inclusion of correct answers in the sampled responses, as well as an advantage reweighting mechanism to strengthen learning from such difficult data samples.
Extensive experiments on the MMAD benchmark demonstrate that EMIT significantly enhances the IAD performance of MLLMs, achieving an average improvement of 7.77\% over the base model (InternVL3-8B) across seven tasks.
% Extensive experiments on the MMAD benchmark demonstrate that EMIT significantly enhances the IAD performance of MLLMs, improving the base model (InternVL3-8B) by an average of 7.77\%  and outperforming the strongest existing MLLM (GLM-4.1V) by 3.03\%.
\end{abstract}

\begin{links}
% \link{Code}{https://anonymous.4open.science/r/EMIT-0F06}
    \link{Code}{https://github.com/guanwei49/EMIT}
    % \link{Datasets}{https://aaai.org/example/datasets}
    % \link{Extended version}{https://aaai.org/example/extended-version}
\end{links}

\section{Introduction}
Industrial anomaly detection (IAD) is essential for ensuring the reliability and safety of modern manufacturing systems \cite{chao2025anomalyr1}. By enabling the timely identification of anomalies, it facilitates prompt intervention, maintenance, and optimization, thereby enhancing the overall efficiency and resilience of industrial production.
Traditional IAD techniques \cite{hu2025dsmbad,li2024memadet,guo2025dinomaly,hoang2025unsupervised,yuan2025mfp} primarily focus on pixel-level or image-level label predictions. 
While effective at detecting the presence of anomalies, they lack the ability to provide comprehensive interpretations or respond to text-based queries.

\begin{figure}[tb]
	% \centering
     \hspace*{-0.1cm} % 向左移动
    \includegraphics[scale=0.35]{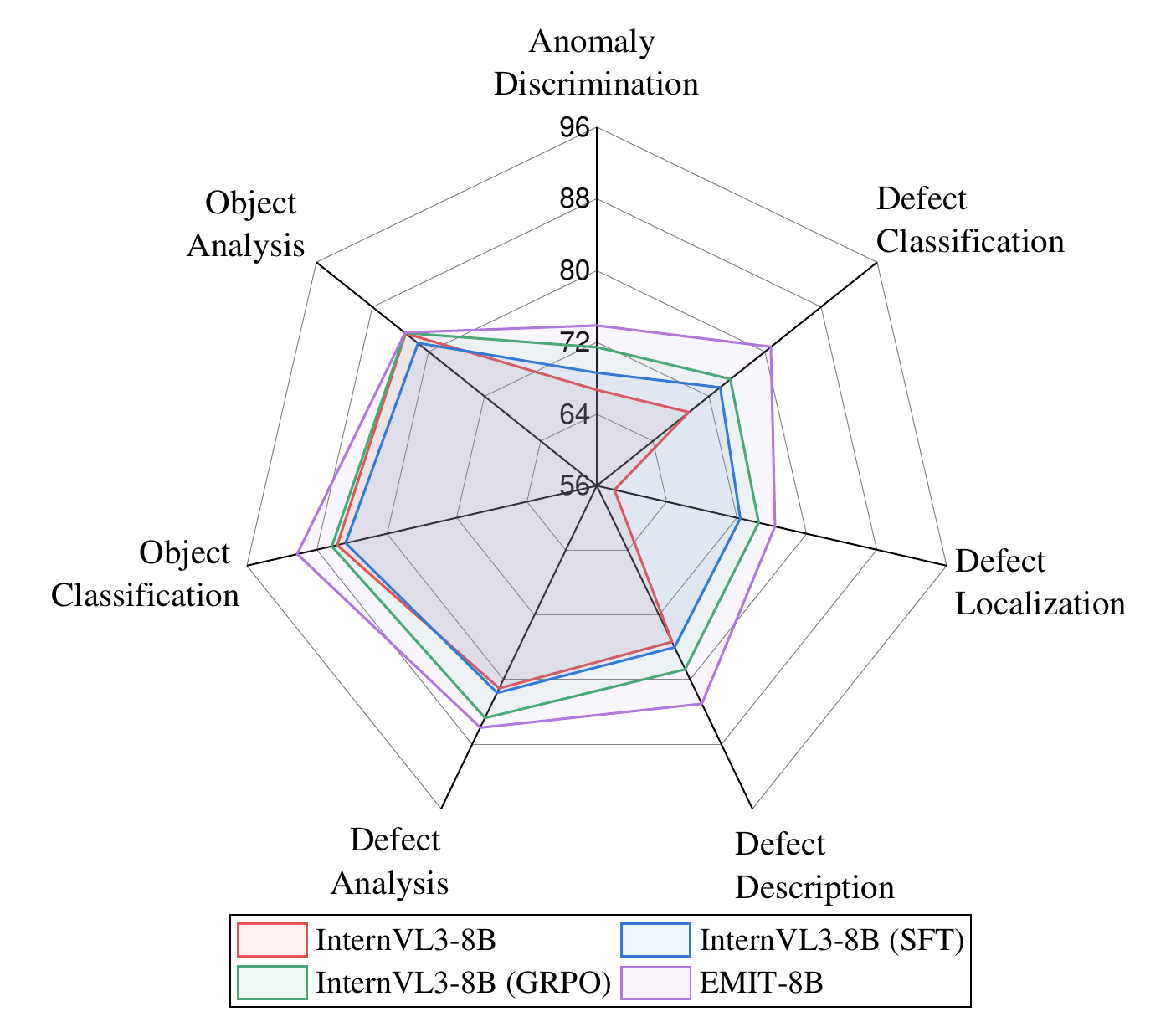}
    	\vskip -0.05in
        \caption{Comparison of models fine-tuned using diverse strategies based on InternVL3-8B.}
	% \caption{Comparison between the strongest MLLM (GLM-4.1V) and InternVL3 under different training strategies with a comparable number of model parameters.}
	\label{fig:comp}
    \vskip -0.15in
\end{figure}

Recently, Multimodal Large Language Models (MLLMs) have witnessed substantial advancements. Models such as Qwen2.5-VL \cite{bai2025qwen25vltechnicalreport}, GLM-4.1V-Thinking \cite{vteam2025glm41vthinkingversatilemultimodalreasoning}, and InternVL3 \cite{zhu2025internvl3exploringadvancedtraining} demonstrate strong multimodal comprehension capabilities and have achieved promising results in tasks such as summarization, paraphrasing, and question answering. 
Some studies \cite{chen2025can, mokhtar2025detect, jin2025logicad, zhang2024logicode} have directly applied these models to IAD without additional fine-tuning. However, as MLLMs are primarily developed for general perceptual and cognitive tasks, their effectiveness in IAD remains limited without domain-specific fine-tuning \cite{xu2025towards}.
Other methods \cite{deng2024vmad,gu2024anomalygpt, li2023myriad,zhang2025eiad,jiang2024fabgpt,xu2025towards} utilize Supervised Fine-Tuning (SFT) to adapt MLLMs for IAD. However, these methods often rely heavily on training data, exhibit limited generalization to real-world anomalies, and optimize fixed training objectives that may not align well with the requirements of downstream tasks \cite{chen2025sft}.
Recent approaches  \cite{li2025lad, zhao2025omniad, zeng2025lr, chao2025anomalyr1}  employ Group Relative Policy Optimization (GRPO) \cite{shao2024deepseekmath} to align MLLMs with downstream tasks via interactive feedback. However, when it comes to challenging data samples, where the MLLM struggles to generate correct answers, plain GRPO may not provide  valid responses, thereby failing to offer effective learning signals.

To address the aforementioned issues,  we propose EMIT, which \textbf{E}nhances \textbf{M}LLMs for \textbf{I}ndustrial anomaly detec\textbf{T}ion via difficulty-aware GRPO.
First, to train our model, we construct a dataset comprising four tasks in IAD, based on existing annotations in publicly available IAD datasets.
In scenarios where defective object images are absent, we leverage GPT-4 \cite{achiam2023gpt} to generate  object descriptive text as a substitute, enabling the model to be trained even without visual defect samples.
For the few-shot anomaly detection scenario, EMIT incorporates a soft prompt and contrastive embeddings into the MLLM. The contrastive embeddings are derived from a heatmap generated via patch-level comparison between the reference and query images.
To train our model, we first freeze the MLLM and align the soft prompt and projector under the SFT framework to enhance anomaly discrimination and defect localization. 
We then fine-tune the entire model using our proposed difficulty-aware GRPO to achieve cohesive and accurate performance.
As shown in Fig. \ref{fig:grpo}, the difficulty-aware GRPO enhances traditional GRPO with two key improvements: a response resampling strategy when no correct response is found, and an  advantage reweighting mechanism that amplifies the advantage for difficult data samples. 
% By incorporating four sophisticated reward functions, our approach is better suited for training on the generated dataset.
As shown in Fig. \ref{fig:comp}, our experiments on the MMAD dataset  demonstrate that our method significantly boosts the performance of the base model (InternVL3-8B) by an average of 7.77\% across seven tasks,  outperforming other finetuning approaches.
% As shown in Fig. \ref{fig: comp}, our experiments on the MMAD dataset  demonstrate that, with a comparable number of model parameters, our method significantly boosts the performance of the base model (InternVL3-8B) by an average of 7.77\% across seven tasks.
% and achieves a 3.03\% average improvement over the existing strongest  MLLM (GLM-4.1V) across seven tasks.
Our contributions are summarized as follows:
\begin{itemize} 
    \item We introduce EMIT, which leverages difficulty-aware GRPO to enhance MLLMs for IAD. Additionally, EMIT incorporates a soft prompt and contrastive embeddings to improve few-shot anomaly detection.
    \item We introduce a difficulty-aware GRPO that incorporates response resampling and advantage reweighting mechanisms.
    \item Extensive experiments show that EMIT achieves the best performance on the MMAD benchmark, significantly enhancing the base model by 7.77\%  across seven tasks.
    % \item Extensive experiments demonstrate that EMIT achieves the best performance on the MMAD benchmark, surpassing the existing best open-source MLLM with a similar parameter volume by an average of 3.03\%.
\end{itemize}

\section{Related Work}
\subsection{Traditional Industrial Anomaly Detection}
Traditional IAD methods can generally be categorized into three main groups. 
Embedding-based methods \cite{hu2025dsmbad,hyun2024reconpatch,jiang2022softpatch,roth2022towards,li2024memadet} rely on extracting embeddings of normal images from a pre-trained vision encoder and storing them in a memory bank, with anomalies detected by comparing the embeddings. 
Reconstruction-based methods \cite{guo2025dinomaly,fan2024revitalizing,fang2023fastrecon,zhang2024realnet,hoang2025unsupervised} train a reconstruction network exclusively on normal samples, enabling anomaly detection based on reconstruction error.
CLIP-based methods \cite{jeong2023winclip, zhou2023anomalyclip,cao2024adaclip,yuan2025mfp,kim2025genclip} explore the generalization capabilities of CLIP and detect anomalies through the alignment of multi-scale image features with text prompts.
Although these methods effectively predict anomalies at the pixel or image level, they fall short in providing comprehensive interpretations or answering text-based questions.

\subsection{MLLMs for Industrial Anomaly Detection}
Driven by the remarkable cognitive capabilities demonstrated by MLLMs, researchers have started exploring their potential applications in IAD.
Several methods \cite{chen2025can, mokhtar2025detect,jin2025logicad,zhang2024logicode} directly apply MLLMs to IAD tasks without fine-tuning.
For instance, Echo \cite{chen2025can} employs four modules to generate specific inputs for MLLMs based on the question type. 
However, since MLLMs are primarily designed for general perceptual and cognitive tasks, their capabilities in IAD are limited without fine-tuning \cite{xu2025towards}.
Other methods \cite{deng2024vmad,gu2024anomalygpt, li2023myriad,zhang2025eiad,jiang2024fabgpt,xu2025towards} employ SFT to adapt MLLMs for IAD. 
For instance, AnomalyGPT \cite{gu2024anomalygpt} adopts two parallel branches to generate anomaly masks and corresponding textual descriptions, trained on synthetic anomalies. Myriad \cite{li2023myriad} utilizes a pre-trained expert model to produce anomaly masks and relies on an MLLM to generate textual responses. Anomaly-OV \cite{xu2025towards} introduces a novel Look Twice Feature Matching (LTFM) mechanism that adaptively selects and emphasizes abnormal visual tokens.
However, SFT-based methods rely heavily on training data, struggle to generalize to real-life anomalies, and optimize fixed objectives that may not align with downstream tasks.
To address this, recent approaches \cite{li2025lad, zhao2025omniad, zeng2025lr, chao2025anomalyr1} adopt plain GRPO to align MLLMs with downstream tasks through interactive feedback. 
Specifically, OmniAD \cite{zhao2025omniad} introduces an initial stage of SFT before applying GRPO.
LR-IAD \cite{zeng2025lr} designs a focal reward mechanism to dynamically emphasize rare answers during training, aiming to address class imbalance.
AnomalyR1 \cite{chao2025anomalyr1} proposes the Reasoned Outcome Alignment Metric (ROAM) as the reward signal, which integrates both the fidelity of the reasoning process and the precision of the final answer.
However, for hard data samples where the sampled response set lacks correct answers, plain GRPO fails to provide effective learning signals, resulting in suboptimal performance.
To address this issue, we propose difficulty-aware GRPO, which introduces a response resampling strategy when no correct response is found and  an  advantage reweighting mechanism that amplifies the advantage for difficult data samples.

\begin{figure*}[tb]
	\centering
    \includegraphics[scale=0.6]{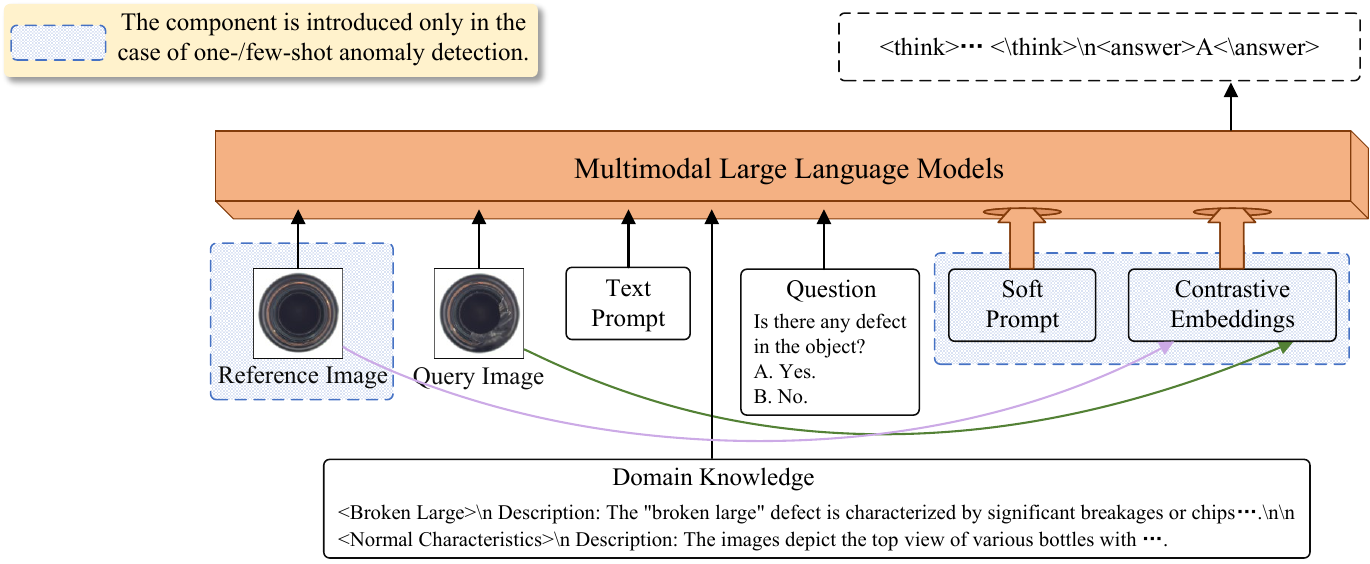}
    	\vskip -0.05in
	\caption{The architecture of EMIT which can integrate components such as the query image, text prompt, domain knowledge, specific question, reference image, soft prompt, and contrastive embeddings. The latter two serve directly as embeddings. Note that the inclusion or exclusion of components is flexible, and  we only present commonly used components in IAD.}
	\label{fig:architecture}
    \vskip -0.15in
\end{figure*}

\section{Methodology}
The architecture of EMIT, as depicted in Fig. \ref{fig:architecture}, is designed to facilitate IAD  through the integration of several crucial components. 
The core input is the query image, which serves as the primary target for analysis. This image is supplemented by a text prompt that introduces the context for the task at hand. For instance, a sample text prompt might read: ``\textit{I have uploaded two images for review, each featuring a product. The product shown in the first image appears to be in perfect condition, free of any defects. To answer a multiple-choice question, please inspect the product in the second image. There is only one correct option.}".
Equally important is the incorporation of domain knowledge, which is instrumental in defining the boundaries of what is considered `normal' versus `anomalous' product.
Another vital component is a specific question that clearly outlines the objective of the analysis.
In the context of one-shot or few-shot anomaly detection scenarios, reference images, typically depicting a normal product, can be utilized alongside the query image. 
The reference image serves as a baseline, facilitating the comparison necessary to identify deviations or anomalies. 
Furthermore, additional inputs such as a soft prompt and  heatmap-guided  contrastive embeddings are constructed to enhance the model's understanding and performance.
Once these inputs are fed into a MLLM, the model processes them to address the posed question and generates a corresponding response.

\begin{figure}[tb]
	\centering
    \includegraphics[scale=0.38]{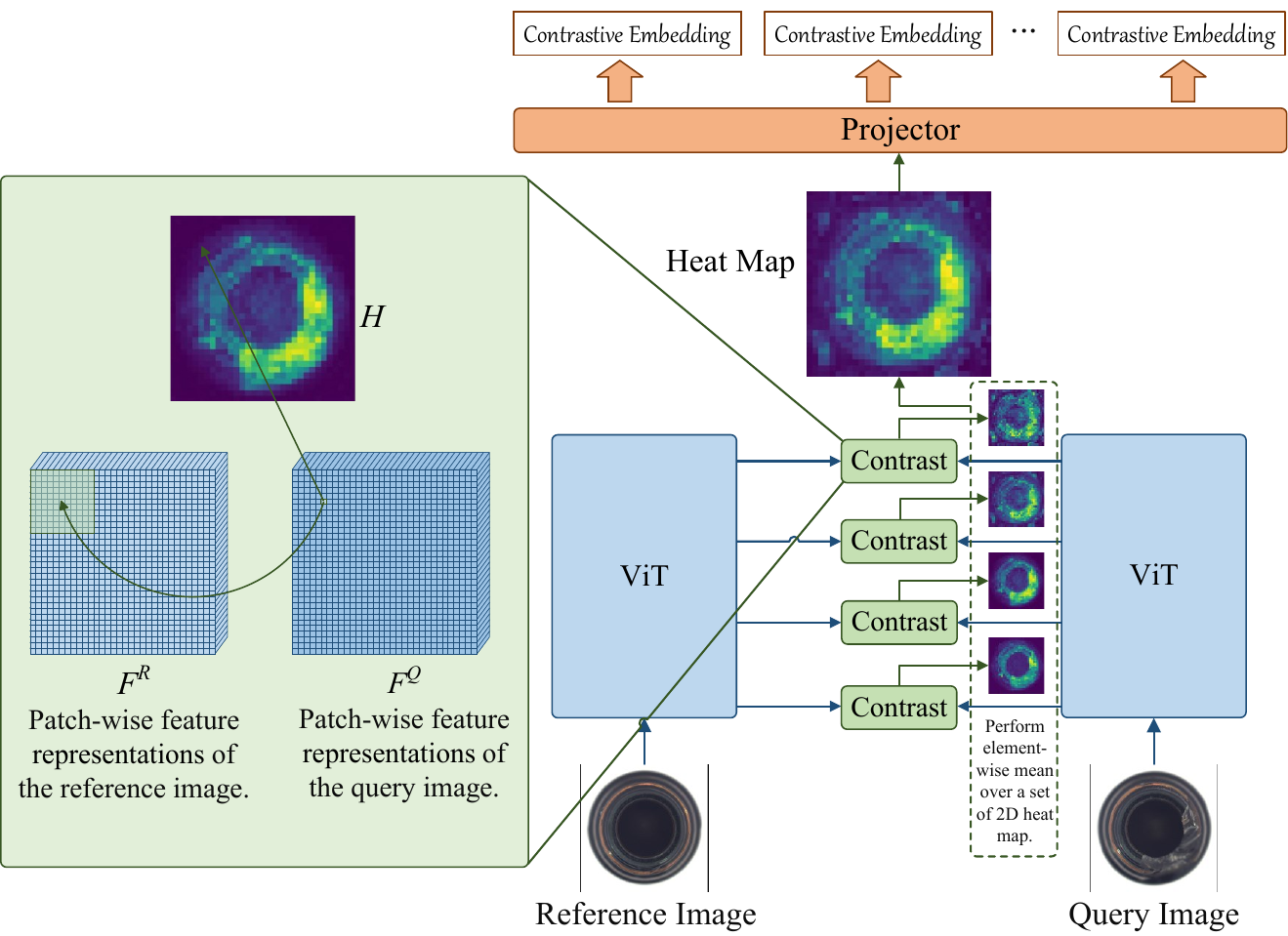}
    	\vskip -0.05in
	\caption{The process for producing contrastive embeddings.}
	\label{fig:contrastmodel}
    \vskip -0.15in
\end{figure}

\begin{figure*}[tb]
	\centering
    \includegraphics[scale=0.65]{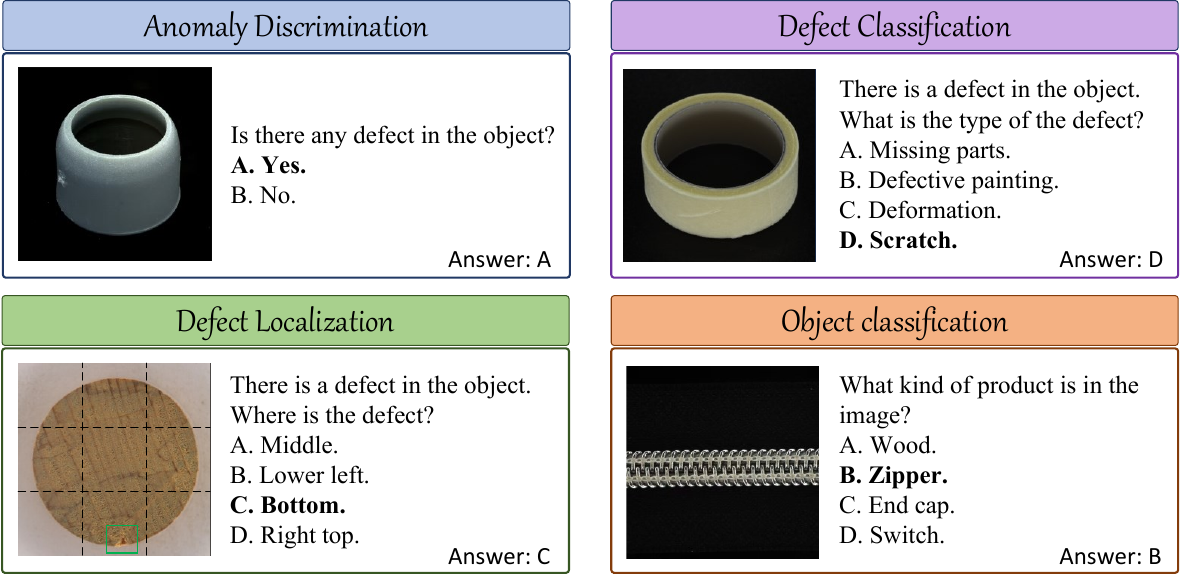}
    	\vskip -0.05in
	\caption{Examples of four tasks from our training dataset.}
	\label{fig:data}
    \vskip -0.15in
\end{figure*}
 
\subsubsection{Contrastive Embeddings}

To help MLLM understand the difference between the reference image and the query image, we generate heatmap-guided contrastive embeddings, as illustrated in Fig.~\ref{fig:contrastmodel}. Specifically, both the reference and query image are processed using a shared ViT of MLLM to extract their respective patch-wise feature representations across multiple transformer layers $l$, denoted as \({}^l F^Q \in \mathbb{R}^{m\times n \times d}\) for the query image and \({}^l F^R\in \mathbb{R}^{m\times n \times d} \) for the reference image, where \(m, n\) are the height and width of   images.
For each transformer layer $l$, a 2D matrix (heat map) ${}^l H\in \mathbb{R}^{m\times n \times 1} $ is generated by determining the minimal cosine distance for every patch in the query image with respect to patches in a local region of the reference image. The detailed process is as follows: for the feature representation of a patch at position \((i, j)\) in the query image at layer \(l\), \( {}^lF_{i,j}^{Q} \), the cosine distance is computed against feature representations of all patches within a defined spatial range \(k\) around the corresponding position in the reference image, denoted as \( {}^lF_{i',j'}^{R} \). The minimal distance is assigned to the heat map at the same position \((i, j)\). Formally,
% Formally, the value of the heat map, \( H_{i,j}^l \), at layer \(l\) is defined as:
\begin{align}
& {}^lH_{i,j} = \min(\{\text{cosine\_dist}({}^lF_{i,j}^{Q}, {}^lF_{i',j'}^{R}) |   i-k \leq i' \leq i+k,  \notag \\  
&  j-k \leq j' \leq j+k, 1 \leq j' \leq n,  1 \leq i' \leq m \})
\end{align}
where \(k\) defines the local range for comparison.
This process is repeated for specific layers of the ViT, producing a set of 2D heat maps, one for each selected layer. Each heat map encodes the difference between the query and reference images at a particular level of feature abstraction, ranging from low-level spatial features to high-level semantic features.

The aggregated heat map reflects the degree and location of defects within the query image. Areas with higher temperatures indicate a greater likelihood of defects. This heat map is subsequently processed through a \textbf{projector}, which maps it into the MLLM embedding space to facilitate deeper analysis. 
The projector consists of a batch normalization layer, a convolutional layer, and a flattening operation. The heat map is first refined by batch normalization and the convolutional layer, capturing spatial relationships. The resulting 2D feature map is then flattened into a sequence of contrastive embeddings to align with the input requirements of the MLLM.
The generated contrastive embeddings empower MLLM to robustly reason about image differences and perform downstream tasks with improved comprehension and accuracy.

\subsubsection{Soft Prompt}
The soft prompt, comprising learnable embeddings fine-tuned during training, serves as a vital intermediary between the specific question being addressed and the contrastive embeddings, ensuring their smooth and coherent integration. 
It is initialized using MLLM embeddings derived from textual cues such as ``Below are some hints for your reference:”. This design substantially bolsters the MLLM's ability to effectively integrate contrastive embeddings, which emphasize visual distinctions. 
% As a result, this approach significantly improves performance in one-shot or few-shot anomaly detection tasks.

\subsection{Dataset Preparation}
Existing defect detection datasets primarily emphasize object class labels, visual mask annotations, and defect class labels. However, they often lack the essential semantic information required for training our model. 

\subsubsection{Task Construction}
To better utilize these datasets, we convert the dataset annotations into four distinct types of multiple-choice questions by applying predefined rules, as demonstrated in Fig. \ref{fig:data}.  This transformation not only enriches the semantic structure of the data but also facilitates the model's training and enhances its overall understanding. The four unique tasks are defined as follows:

\begin{itemize}
    \item \textit{Anomaly Discrimination:} This is a binary classification task where the model determines whether a defect exists in the query image. 
    % The goal is to enhance the model's ability to detect the presence of anomalies.
    \item
    \textit{Defect Classification:} This task requires the model to classify the type of defect, serving as a test of its semantic understanding of various industrial anomalies.
    \item
    \textit{Defect Localization:} This task challenges the model to identify the approximate location of defects. Each image is divided into a $3 \times 3$ grid of nine equal regions, referred to as `top left', `top right', etc. The ground truth answer can be determined by transforming visual mask annotations according to predefined rules.
    \item
    \textit{Object Classification:} This task focuses on the categorization of industrial products depicted  in query images. 
    % By understanding normal and defective object characteristics, the model is better equipped to recognize anomalies effectively.
\end{itemize}

\begin{figure}[ht]
    	\centering
    	% \vskip -0.15in
    \subfloat[Objects `\textit{object\_type}' with defect `\textit{defect\_type}'.]{
	\centering
	\includegraphics[scale=0.64]{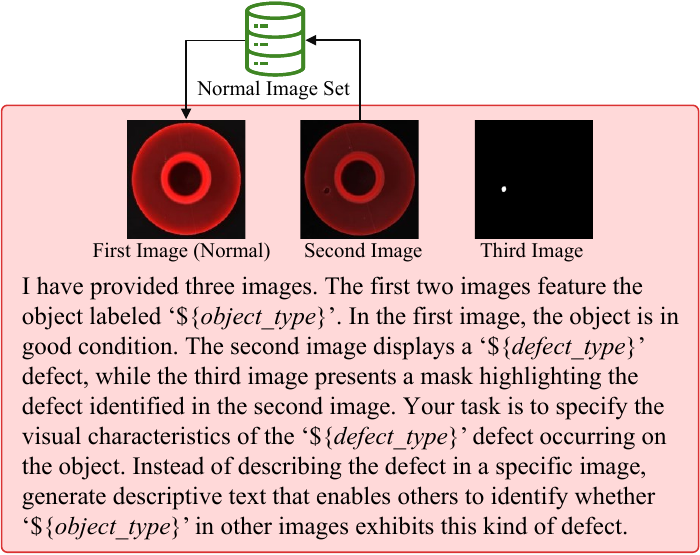}
	\label{fig:domain_knowledge_gen_defect}
	}\\ 	
        \vspace{-5pt}
% 	\hfil
\subfloat[Normal objects `\textit{object\_type}'.]{
	\centering
	\includegraphics[scale=0.64]{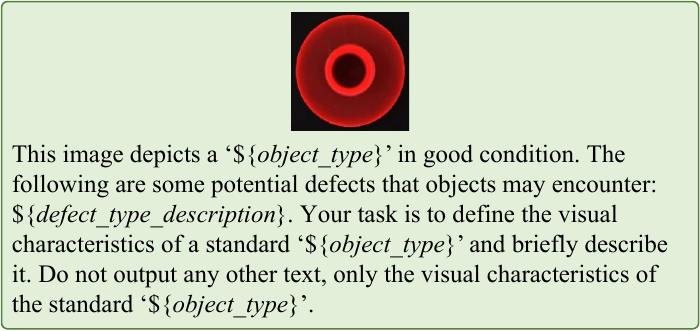}
	\label{fig:domain_knowledge_gen_normal}
	}
	\vskip -0.05in
	\caption{The prompt for constructing domain knowledge.}
	\vskip -0.2in
\end{figure}

% \begin{figure}[htb]
% 	\centering
% \includegraphics[scale=0.7]{pic/domain_knowledge_gen_defect.pdf}
%     	\vskip -0.05in
% 	\caption{The prompt for defining characteristics of objects `\textit{object\_type}' with defect `\textit{defect\_type}'.}
% 	\label{fig:domain_knowledge_gen_defect}
%     \vskip -0.15in
% \end{figure}
% \begin{figure}[htb]
% 	\centering
% 	\includegraphics[scale=0.7]{pic/domain_knowledge_gen_normal.pdf}
%     	\vskip -0.05in
% 	\caption{The prompt for defining characteristics of normal objects `\textit{object\_type}'.}
% 	\label{fig:domain_knowledge_gen_normal}
%     \vskip -0.15in
% \end{figure}

\subsubsection{Domain Knowledge Construction}
% clip-vit-base-patch32用于检索相似图
To construct domain knowledge, i.e.,  the visual characteristics of normal objects and anomalies, we utilize GPT-4 Turbo\footnote{https://platform.openai.com/docs/models/gpt-4-turbo}.

\textit{Defining defect characteristics}:
To define the visual characteristics of objects  `\textit{object\_type}' exhibiting defects  `\textit{defect\_type}', we provide GPT with three images: i) an image of the defective object, ii) a corresponding mask highlighting the location of the defect, and iii) a similar image of a normal object of the same type, retrieved from the normal image set.
The prompt then instructs GPT to generate a detailed and generalized description of the defect’s visual characteristics based on the given images.
The specific prompt used for this task is shown in Fig.~\ref{fig:domain_knowledge_gen_defect}.  

\textit{Defining normal characteristics}:
To describe the typical visual characteristics of normal objects classified as `\textit{object\_type}', we provide GPT with an image of a normal object of type `\textit{object\_type}' along with contextual information about potential defect characteristics.
Using these inputs, GPT is instructed to produce a thorough and generalized description of the normal object’s visual characteristics. 
The specific prompt for this task is shown in Fig.~\ref{fig:domain_knowledge_gen_normal}.

\begin{figure}[t]
    	\centering
    	% \vskip -0.15in
    \subfloat[An anomalous object with defect `\textit{defect\_type}' at location `\textit{defect\_location}'.]{
	\centering
	\includegraphics[scale=0.64]{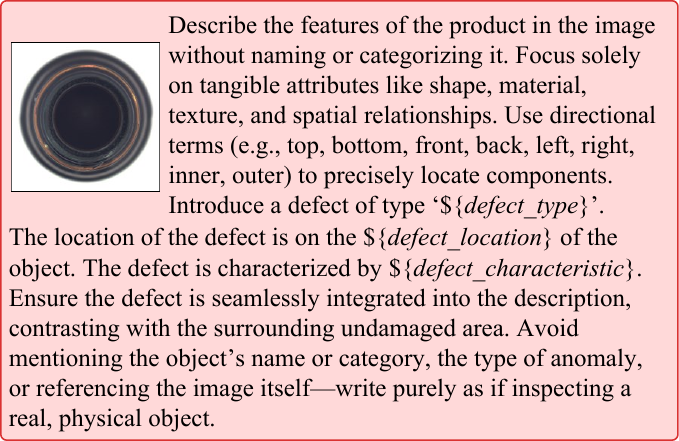}
	\label{fig:textdata_defect}
	}\\
    	% \vskip -0.01in
        \vspace{-5pt}
% 	\hfil
\subfloat[Normal objects `\textit{object\_type}'.]{
	\centering
	\includegraphics[scale=0.64]{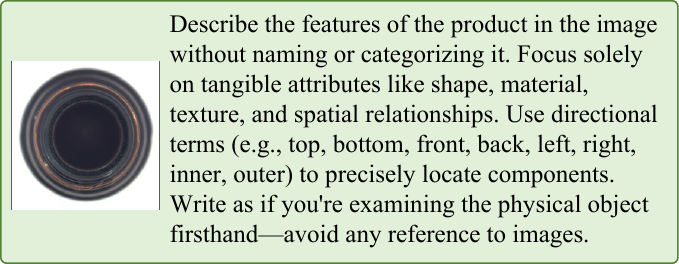}
\label{fig:textdata_normal}
	}
	\vskip -0.05in
	\caption{The prompt for generating descriptive text.}
	\vskip -0.15in
\end{figure}

% \begin{figure}[tb]
% 	\centering
% 	\includegraphics[scale=0.7]{pic/textdata_defect.pdf}
%     	\vskip -0.05in
% 	\caption{The prompt for generating descriptive text for an anomalous object with defect `\textit{defect\_type}' at location `\textit{defect\_location}'.}
% 	\label{fig:textdata_defect}
%     \vskip -0.15in
% \end{figure}
% \begin{figure}[tb]
% 	\centering
% \includegraphics[scale=0.7]{pic/textdata_normal.pdf}
%     	\vskip -0.05in
% 	\caption{The prompt for generating descriptive text of a normal object.}
% \label{fig:textdata_normal}
%     \vskip -0.15in
% \end{figure}

\subsubsection{Object Descriptive Text as Query Image}
For scenarios where only images of normal objects and expert domain knowledge, specifically detailing the defect type and its definition, are available, it becomes challenging to directly utilize this information to train the model due to the absence of defective object images.
However, thanks to the semantic alignment between images and text in MLLMs, when the image and text consistently refer to the same object or concept, the model is more likely to provide consistent answers to questions.
This capability allows us to shift the reliance from defective object images to textual descriptions of the defective objects.
This approach eliminates the uncertainties and high costs associated with fabricating defective object images.

With the assistance of GPT, we can input an image of a normal object, specify the `\textit{defect\_type}' and `\textit{defect\_location}', and instruct GPT to generate descriptive text that details the visual characteristics of the object as if it exhibited a defect of type `\textit{defect\_type}' at location `\textit{defect\_location}'. The specific prompt used for this process is shown in Fig.~\ref{fig:textdata_defect}.
Similarly, we can input an image of a normal object and instruct GPT to generate descriptive text capturing the visual characteristics of the object. The specific prompt for this process is shown in Fig.~\ref{fig:textdata_normal}.

\subsubsection{Data Collection}
\begin{figure}[tb] 
       \hspace*{-0.3cm} % 向左移动
    \includegraphics[scale=0.43]{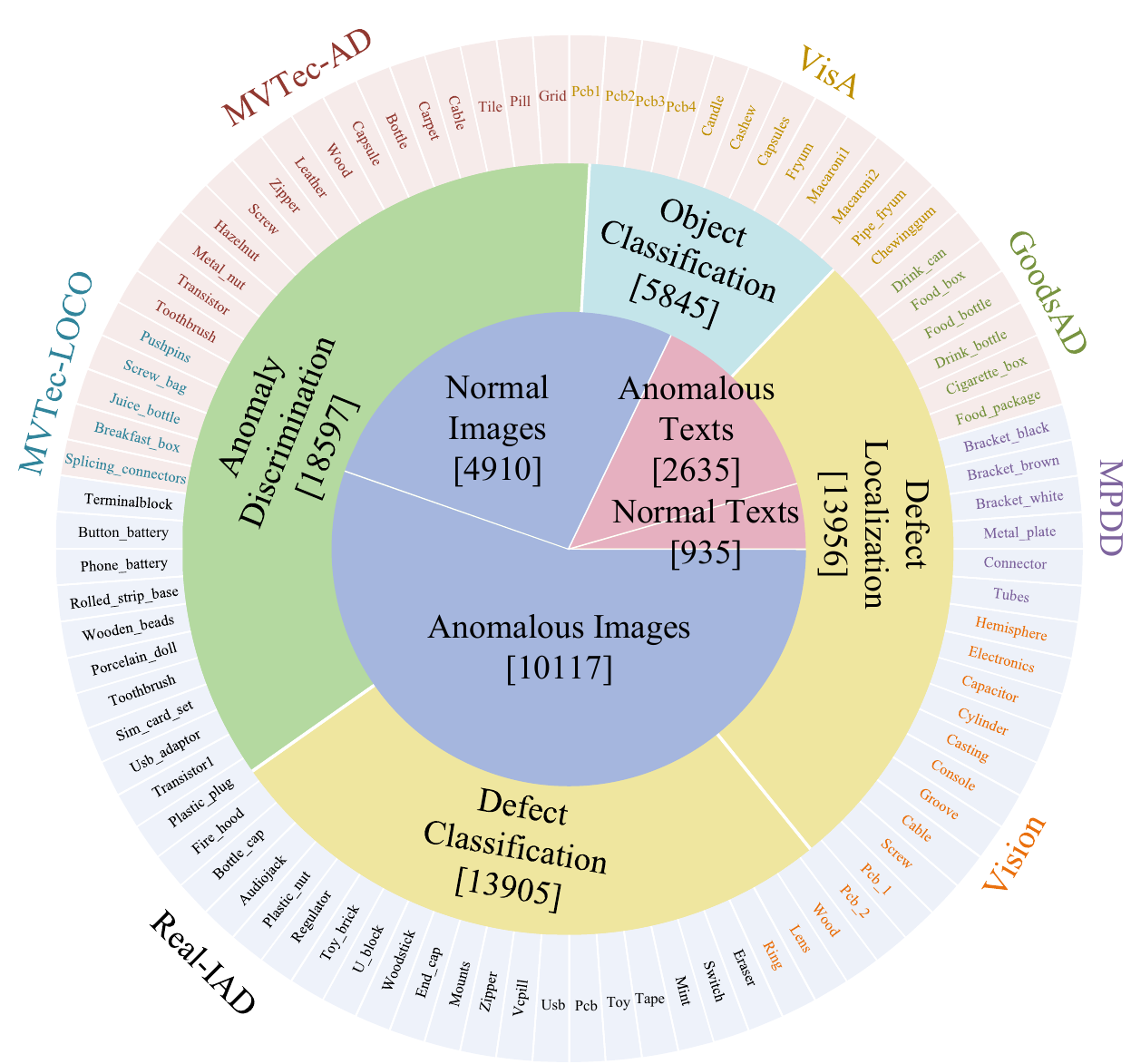}
    	\vskip -0.05in
	\caption{Statistical overview of the training dataset. The innermost layer represents image components, the middle layer depicts subtask composition, and the outermost layer illustrates object categories.}
	\label{fig:data_sta}
    \vskip -0.15in
\end{figure}

We utilize three industrial anomaly detection datasets, Vision \cite{bai2023vision}, Real-IAD \cite{wang2024real}, and MPDD \cite{jezek2021deep}, to directly construct four distinct tasks. 
Since our evaluation is based on the MMAD dataset \cite{jiang2024mmad}, we carefully handle its subdatasets, including MVTec AD \cite{bergmann2019mvtec}, MVTec LOCO AD \cite{bergmann2022beyond}, VisA \cite{zou2022spot}, and GoodsAD \cite{zhang2024pku}, to prevent data leakage. 
Specifically, instead of directly using these subdatasets to create four tasks, we randomly select five images of normal objects for each object type and generate descriptive text for these images to serve as query images. 
Finally, statistical information pertaining to the training dataset is presented in Fig. \ref{fig:data_sta}.

\begin{figure*}[tb]
	\centering
    \includegraphics[scale=0.5]{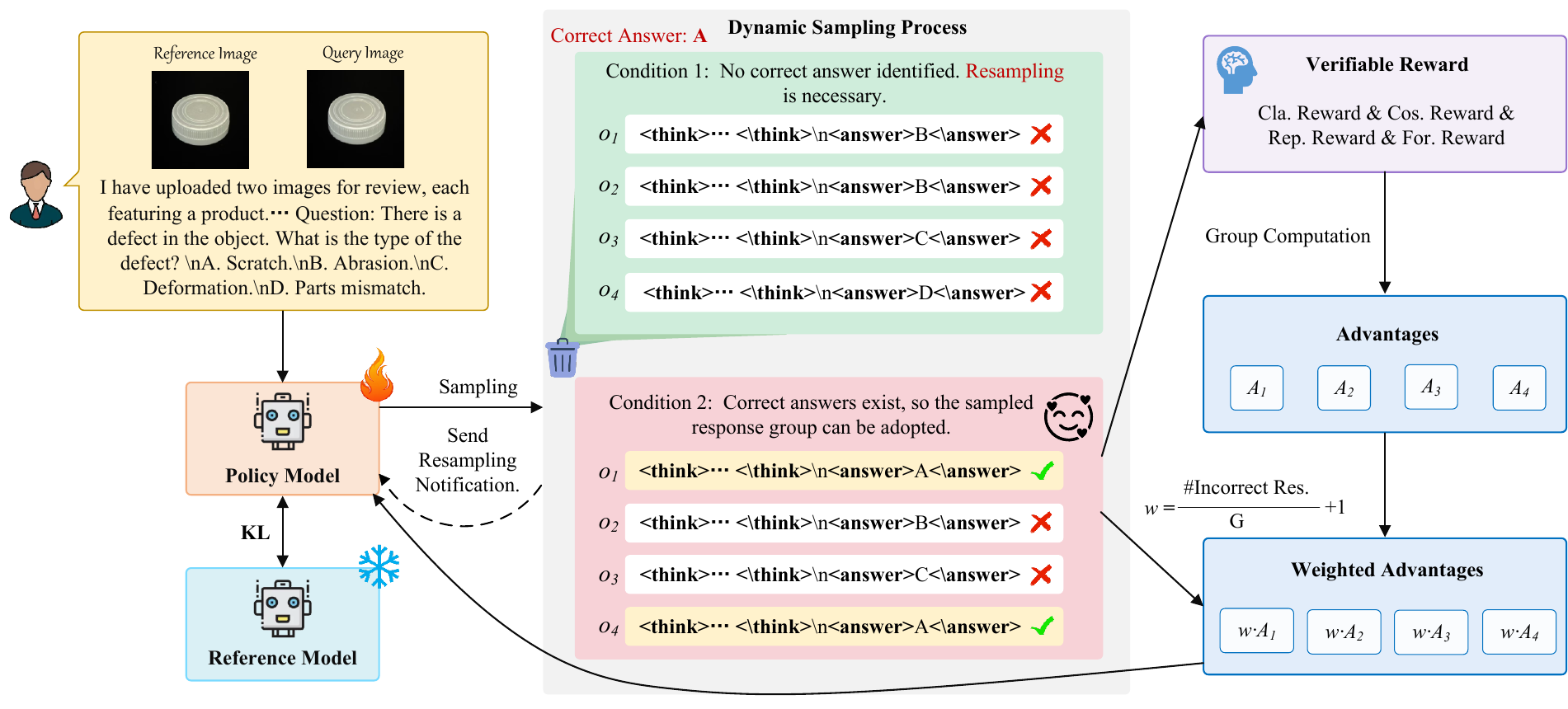}
    	\vskip -0.05in
	\caption{The framework of difficulty-aware GRPO for IAD.}
	\label{fig:grpo}
    \vskip -0.15in
\end{figure*}

\subsection{Difficulty-Aware GRPO for IAD}
Original GRPO enhances the Proximal Policy Optimization (PPO) \cite{schulman2017proximal}  by focusing on group-based evaluation of advantages, thereby bypassing the need for a value function. 
In detail, for each question-answer pair $(q, a)$, the policy model $\pi_{\theta_{\text{old}}}$ generates a collection of $G$ responses, denoted as $\{o_i\}_{i=1}^G$. The advantage estimate for the $i$-th response is then determined by standardizing the associated rewards $R$ across the group:

\begin{equation}
    \hat{A}_{i, t} = \frac{r_i - \text{mean}(\{R_j\}_{j=1}^G)}{\text{std}(\{R_j\}_{j=1}^G)}
\end{equation}

% GRPO incorporates a clipped surrogate objective, akin to that of PPO. However, 
% GRPO computes loss at the granularity of individual samples within each sequence, and only later aggregates the average loss across sequences. This approach can unintentionally cause tokens within longer responses (which contain more tokens) to have a disproportionately smaller contribution to the overall loss. 
We adopt the token-level policy gradient loss proposed in \cite{yu2025dapo}, augmented by the addition of a KL divergence penalty. 
% This approach ensures the policy remains close to a reference policy while providing a more uniform weighting across tokens. 
The training objective is formally defined as follows:

\begin{align}
    \mathcal{J}(\theta) = & 
    \mathbb{E}_{(q, a) \sim \mathcal{D}, \{o_i\}_{i=1}^G \sim \pi_{\theta_{\text{old}}}(\cdot | q)}  \\
    & \Bigg[ 
        \frac{1}{\sum_{i=1}^{G}|o_i|} \sum_{i=1}^G \sum_{t=1}^{|o_i|} \Big( M_{i,t} \notag  
        - \beta D_{\text{KL}} \left( \pi_\theta || \pi_{\text{ref}} \right) \Big) 
    \Bigg] \label{eq:grpo_objective}
\end{align}
where
\begin{align}
    M_{i,t} = & \min \Bigg[ 
        \frac{\pi_\theta(o_{i,t} | q, o_{i,<t})}{\pi_{\theta_{\text{old}}}(o_{i,t} | q, o_{i,<t})} \hat{A}_{i,t}, \notag \\
        & \text{clip} \Bigg( \frac{\pi_\theta(o_{i,t} | q, o_{i,<t})}{\pi_{\theta_{\text{old}}}(o_{i,t} | q, o_{i,<t})}, 1 - \epsilon, 1 + \epsilon \Bigg) \hat{A}_{i,t} 
    \Bigg] 
\end{align}

As shown in Fig. \ref{fig:grpo}, to better handle difficult data samples, i.e., cases where the MLLM struggles to generate correct answers, we propose a difficulty-aware GRPO that extends the original GRPO by incorporating two core innovations: \textit{response resampling} and \textit{advantage reweighting}. 

\subsubsection{Response Resampling}
For a given question, if none of the responses in the sampled set $\{o_i\}_{i=1}^G$ contains a correct answer, we re-engage the policy model to generate a new collection of $G$ responses. This ensures that at least one response includes the correct answer, thereby enabling the model to consistently learn from valid supervision signals. Without such a mechanism, the model may be misled by low-quality, entirely incorrect responses, resulting in degraded training quality. 

\subsubsection{Advantage Reweighting}
To make hard questions receive more attention during training, we adjust the advantage based on the error distribution in the sampled responses. Specifically, the proportion of responses containing incorrect answers to the total number of sampled responses is used to measure question difficulty: the greater the proportion of incorrect responses  (determined by comparing the selected choice in the `answer' tag to the ground truth) the harder the question.

For each question, a  weight \( w \) is applied to the advantage values of all responses. The weight \( w \) is defined as:
\begin{equation}
    w = \frac{\text{\# Incorrect Responses}}{G} + 1
\end{equation}
where \( G \) is the total number of sampled responses. The addition of \( 1 \) ensures that the weight does not become zero, allowing the model to still learn from even the easiest questions (all responses are correct).
This weight \( w \) is then used to scale the original advantage value, \( \hat{A}_{i,t} \), yielding the reweighted advantage:
\begin{equation}
    \hat{A}_{i,t}^\text{reweighted} = w \cdot \hat{A}_{i,t}
\end{equation}
By applying this weight, the absolute value of the advantage for harder questions becomes larger, which ensures that harder questions receive more attention during training, improving the model's learning from them.

\begin{table*}[]
	\caption{Accuracy (\%) over the MMAD dataset under one-shot anomaly detection.}
      \vskip -0.1in 
 \label{tab:comp}
 \footnotesize
    \centering
    % \resizebox{\linewidth}{!}{
      \setlength{\tabcolsep}{1mm}{
   \begin{tabular}{c|c|c|cccc|cc|c}
   \toprule
      \multirow{2}{*}{Model}                &    \multirow{2}{*}{Scale}    & Anomaly        & \multicolumn{4}{c|}{Defect}                             & \multicolumn{2}{c|}{Object} &      \multirow{2}{*}{Average}      \\
    \cmidrule(lr){3-3}  \cmidrule(lr){4-7}  \cmidrule(lr){8-9} 
                      &     & Discrimination & Classification & Localization & Description & Analysis & Classification  & Analysis &   \\  \midrule
Random Choice         &   -  & 50.00                              & 25.00                              & 25.00                            & 25.00                           & 25.00                        & 25.00                              & 25.00                        & 28.57                       \\ \midrule
LLaVA-OneVision       & 7B  & 55.53                              & 71.18                              & 55.21                            & 74.24                           & 80.53                        & 90.53                              & 82.77                        & 72.86                       \\
LLaVA-NeXT-Interleave & 7B  & 58.25                              & 64.58                              & 51.48                            & 66.01                           & 76.93                        & 87.71                              & 76.07                        & 68.72                       \\
MiMo-VL               & 7B  & 56.39                              & 64.78                              & 62.71                            & 75.06                           & 72.60                        & 89.56                              & 81.38                        & 71.78                       \\
Qwen2.5-VL            & 7B  & 64.42                              & 69.35                              & 61.49                            & 74.65                           & 81.57                        & 92.46                              & 85.96                        & 75.70                       \\
GLM-4.1V-Thinking     & 9B  & 72.31                              & 72.01                              & 66.47                            & 79.98                           & 84.61                        & \textbf{93.79}                              & 83.26                        & 78.92                       \\
Pixtral               & 12B & 56.61                              & 64.92                              & 54.44                            & 72.50                           & 82.29                        & 85.74                              & 77.54                        & 70.58                       \\
Gemma-3               & 12B & 51.60                              & 64.98                              & 49.75                            & 66.81                           & 78.63                        & 89.43                              & 77.30                        & 68.36                       \\
Kimi-VL               & 16B & 66.57                              & 71.72                              & 61.88                            & 79.78                           & 82.93                        & 92.18                              & \textbf{86.51}                        & 77.37                       \\
AnomalyR1               & 3B & 53.47 &	69.95 &	64.99 	& 77.27 &	81.89 & 	89.71 &	85.19 &	74.64  \\
InternVL3 (base)   & 8B  & 66.71                              & 69.16                              & 58.01                            & 75.33                           & 81.06                        & 85.68                              & 83.28                        & 74.18                       \\
InternVL3 (SFT)    & 8B  & 68.59                              & 73.59                              & 72.42                            & 76.00                           & 81.65                        & 84.71                              & 81.52                        & 76.92                       \\
InternVL3 (GRPO)   & 8B  & 71.44                              & 75.07                              & 74.51                            & 78.72                           & 84.73                        & 86.32                              & 83.35                        & 79.16                       \\
EMIT                  & 8B  & \textbf{73.87}                              & \textbf{80.85}                              & \textbf{76.39}                            & \textbf{83.00}                           & \textbf{85.92}                        & 90.26                              & 83.37                        & \textbf{81.95}              
 \\ 
 \bottomrule
    \end{tabular}
    }
    % }
    \vskip -0.1in 
\end{table*}

\subsubsection{Verifiable Rewards}
% The reward function we use consists of four components: format reward $R_{\text{For.}}$, classification accuracy reward $R_{\text{Cla.}}$, cosine reward $R_{\text{Cos.}}$ and repetition reward $R_{\text{Rep.}}$.
% The reward function we use consists of four components: \textit{format reward}, \textit{classification accuracy reward}, \textit{cosine reward} and \textit{repetition reward}.

% The format reward guides the model to structure its output with the reasoning process enclosed in the `<think>' tag and the final answer in the `<answer>' tag. A reward value of 1 is granted for strict adherence to this format, while a reward value of 0 is applied for omissions or incorrect usage of these tags.
% The classification reward and cosine reward is calculated based on the answer in the `<answer>` tag. Considering our training data is the multiple choice question,  
% If the generated answer is both a valid choice and correct, the classification reward is 1, and cosine reward encourage the model output shorter think process.
% If it is  a valid choice but is incorrect, the classification reward is 0 and cosine reward encourage the model output larger think process.
% If the answer does not correspond to any of the provided choices,  the classification reward and cosine reward is -1.
% the repetition reward penalizes repetitive patterns in the generated text  based on n-gram, with a penalty proportional to high repetition levels.

% Finally, the total reward  is computed by  This weighted sum,which ensures that each aspect of quality is appropriately balanced.  

The reward function consists of four components: \textit{format reward}, \textit{classification accuracy reward}, \textit{cosine reward} \cite{yeo2025demystifyinglongchainofthoughtreasoning}, and \textit{repetition reward}.

The \textit{format reward} guides the model to structure its output with the reasoning process enclosed in the `think' tag and the final answer in the `answer' tag. A reward value of 1 is granted for strict adherence to this format, while a reward value of 0 is applied for omissions or incorrect usage of these tags.  
The \textit{classification reward} and \textit{cosine reward} are calculated based on the answer provided within the `answer' tag. Considering the fact that our training data comprises multiple-choice questions: if the generated answer is a valid choice (i.e., \texttt{A}, \texttt{B}, etc.) and correct, the classification reward is 1, and the cosine reward encourages the model to output a shorter reasoning process within the `think' tag.
If the answer is a valid choice but incorrect, the classification reward is 0, and the cosine reward encourages the model to produce a more detailed and extensive reasoning process.
If the answer does not correspond to any of the provided choices, both the classification reward and cosine reward are set to -1, discouraging invalid answers.
The \textit{repetition reward} penalizes repetitive patterns in the generated text, based on the frequency of repeated n-grams. The penalty is proportional to the level of repetition, with higher rewards  assigned to responses containing fewer repeated n-grams.
Finally, the total reward is computed as a weighted sum of these components. 
% This approach ensures that different aspects of output quality, such as reasoning structure, correctness, brevity, and diversity, are appropriately balanced during model optimization.

\subsection{Training Strategy} 
% The training of EMIT consists of two main stages:

% \begin{itemize}
%     \item   \textit{Stage 1:} In this stage, the MLLM is kept frozen, while we focus on aligning the soft prompt and projector to the MLLM. This alignment ensures that the soft prompt  and contrastive embeddings effectively support the MLLM in achieving enhanced performance in anomaly discrimination and defect localization tasks, leveraging the SFT schema.
    
%   \item  \textit{Stage 2:} This stage involves fine-tuning the entire model using our proposed difficulty-aware GRPO  for cohesive and accurate performance. 
% \end{itemize}

To train EMIT, we follow two main stages: 
i) In the \textit{first} stage, the MLLM is kept frozen, while we focus on aligning the soft prompt and projector to the MLLM. This alignment ensures that the soft prompt  and contrastive embeddings effectively support the MLLM in achieving enhanced performance in anomaly discrimination and defect localization tasks, leveraging the SFT schema.
ii) In the \textit{second} stage, we fine-tune the entire model using our proposed difficulty-aware GRPO for cohesive and accurate performance.

\section{Experiments}
\subsection{Experimental Setup}
\subsubsection{Test Datasets}
We use the MMAD dataset \cite{jiang2024mmad} as our evaluation benchmark. This dataset includes seven subtasks: anomaly discrimination, defect classification, defect localization, defect description, defect analysis, object classification, and object analysis. Additionally, it provides domain knowledge for each object type, making it a comprehensive resource for evaluation.

\subsubsection{Compared Methods}
We compare our fine-tuned model with nine state-of-the-art, general-purpose, open-source MLLMs, including LLaVA-OneVision \cite{li2024llavaonevisioneasyvisualtask}, LLaVA-NeXT-Interleave \cite{li2024llavanextinterleavetacklingmultiimagevideo}, MiMo-VL \cite{coreteam2025mimovltechnicalreport}, Qwen2.5-VL \cite{bai2025qwen25vltechnicalreport},  GLM-4.1V-Thinking \cite{vteam2025glm41vthinkingversatilemultimodalreasoning}, Pixtral \cite{agrawal2024pixtral12b}, Gemma 3 \cite{gemmateam2025gemma3technicalreport}, Kimi-VL \cite{kimiteam2025kimivltechnicalreport} and InternVL3 \cite{zhu2025internvl3exploringadvancedtraining}.
In addition, we include AnomalyR1 \cite{chao2025anomalyr1}, a variant of Qwen2.5-VL specifically fine-tuned using GRPO for the IAD task. We directly use the model parameters released by the authors.

% \subsubsection{Evaluation Metrics}
% Following existing anomaly detection methods, we employ \textit{precision}, \textit{recall}, \textit{F}$_1$-\textit{score} and \textit{accuracy} to evaluate the anomaly detection performance. 
% The recall-oriented understudy for gisting evaluation (ROUGE) \cite{lin2004rouge} is a software package and metric set designed to assess the quality of generated text by comparing it with ground truth text. In our evaluation of DABL's ability to interpret the cause of anomalies, we utilize \textit{ROUGE-2} and \textit{ROUGE-L} metrics.
% We conduct each experiment five times and report the mean results.

\begin{table*}[h]
	\caption{Ablation study.  `\textit{w/o}' denotes `\textit{without}'}
      \vskip -0.1in 
 \label{tab:ab}
 \footnotesize
    \centering
    % \resizebox{\linewidth}{!}{
      \setlength{\tabcolsep}{1mm}{
   \begin{tabular}{c|c|cccc|cc|c}
   \toprule
      \multirow{2}{*}{Model}                  & Anomaly        & \multicolumn{4}{c|}{Defect}                             & \multicolumn{2}{c|}{Object} &      \multirow{2}{*}{Average}      \\
    \cmidrule(lr){2-2}  \cmidrule(lr){3-6}  \cmidrule(lr){7-8} 
         & Discrimination & Classification & Localization & Description & Analysis & Classification  & Analysis &   \\  \midrule
w/o Stage1                            & 72.49                              & 79.83                              & 75.88                            & 82.14                           & 85.21                        & 88.53                              & 83.35                        & 81.06                       \\
w/o Stage2                            & 67.24                              & 69.16                              & 65.58                            & 75.33                           & 81.06                        & 85.68                              & 83.28                        & 75.33                       \\
w/o Text Samples                       & 71.07                              & 76.74                              & 75.98                            & 80.36                           & 82.55                        & 87.12                              & 81.76                        & 79.37                       \\
 EMIT              & \textbf{73.87}                              & \textbf{80.85}                              & \textbf{76.39}                            & \textbf{83.00}                           & \textbf{85.92}                        & \textbf{90.26}                              & \textbf{83.37}           & \textbf{81.95}  \\ 
 \bottomrule
    \end{tabular}
    }
    % }
    \vskip -0.1in 
\end{table*}

\subsubsection{Implementation Details}
We config open-source InternVL3 8B\footnote{https://huggingface.co/OpenGVLab/InternVL3-8B} as our base MLLM,  enhancing its capability in IAD. 
For training stage 2, referred to as GRPO, we set the weight of KL divergence $\beta$ in loss function to 0.01 and sample $G=8$ responses during training. 
Regarding the reward function, we assign a weight of $3$ to the classification accuracy reward and a weight of $1$ to all other reward components.
To address the computational cost associated with fine-tuning MLLMs with a large parameter count, we adopt QLoRA \cite{dettmers2024qlora} to reduce memory usage efficiently. Specifically, we set $lora\_rank$ to $8$ and $lora\_alpha$ to $32$.
We train the model using the AdamW optimizer \cite{loshchilov2017decoupled} for both stages. The training process spans two epochs for stage $1$ and one epoch for stage $2$. The learning rate is $1 \times 10^{-5}$, with a warm-up phase followed by cosine decay. The mini-batch size is set to $64$ throughout. The training is carried out on eight NVIDIA A100 GPUs, each with 80 GB of memory.

During training and testing, we adopt a default one-shot setting. In this setup, a reference image is provided alongside the query image. The reference image is selected as the most similar image from the normal image set, identified using the \textit{clip-vit-base-patch32} model\footnote{https://huggingface.co/openai/clip-vit-base-patch32}.
For tasks related to defects, such as anomaly discrimination, defect classification, defect localization, defect description, and defect analysis, we also incorporate domain knowledge. However, for tasks such as object classification and object analysis, providing domain knowledge could potentially reveal the object type, and hence, it is omitted in these cases. 

The soft prompt and contrastive embeddings are provided exclusively for tasks such as anomaly discrimination and defect localization. Additionally, during training, when only the object descriptive text is provided instead of the query image, we omit the use of the soft prompt and contrastive embeddings. This choice is driven by the requirement that generating contrastive embeddings necessitates both a query image and at least one reference image.
 
\subsubsection{Metrics}
Aligned with the MMAD benchmark, which comprises multiple-choice problems, we utilize \textit{accuracy} as our evaluation metric. This is calculated by dividing the number of correct predictions by the total number of predictions.

\subsection{Comparative Results}
Table \ref{tab:comp} showcases the performance of EMIT in comparison to several MLLMs on the MMAD dataset. EMIT demonstrates exceptional results on the MMAD benchmark, surpassing the best-performing open-source MLLM, GLM-4.1V-Thinking-9B, which has a similar parameter size, with an average improvement of 3.03\%.
Notably, EMIT significantly enhances the IAD performance of the base model, InternVL3-8B, achieving an average improvement of 7.77\% across seven tasks. 
This highlights the effectiveness of our approach in advancing the capabilities of MLLMs for IAD-related tasks.
Compared to the domain-specific method AnomalyR1, which fine-tunes Qwen2.5-VL using GRPO, EMIT achieves an average improvement of 7.31\%, benefiting from a well-curated and comprehensive training dataset, contrastive embeddings, and the difficulty-aware GRPO.
Moreover, EMIT consistently outperforms other MLLMs on defect-related tasks, including anomaly discrimination, defect classification, defect localization, defect description, and defect analysis.
For object-related tasks such as object classification and analysis, EMIT achieves slightly lower accuracy than the best-performing MLLM, primarily due to the limitations of its base model (InternVL3). However, compared to its base model, EMIT shows significant improvements in object classification. As for object analysis, only marginal gains are observed, likely because the training dataset lacks relevant data samples for this task.

\subsection{Ablation Study}
\subsubsection{Training Strategy}
% the seconde stage is replaced by traditional grpo, sft.
To further evaluate the benefits of our difficulty-aware GRPO, we compare it with standard GRPO and SFT by substituting these methods for training Stage 2. 
The results, shown in Table \ref{tab:comp}, reveal that both the standard GRPO and our difficulty-aware GRPO lead to performance improvements across all tasks. In contrast, SFT degrades the base model's performance on object classification and analysis by 0.97\% and 1.76\%, respectively.
Furthermore, training with our difficulty-aware GRPO yields, on average, a 5.03\% higher accuracy compared to SFT and a 2.79\% higher accuracy compared to standard GRPO.  
% The results, shown in Table \ref{tab:comp}, reveal that training with our difficulty-aware GRPO yields, on average, a 5.03\% higher accuracy compared to SFT and a 2.79\% higher accuracy compared to standard GRPO. 
These findings underscore the limitations of SFT in terms of generalization, as well as the inability of traditional GRPO to effectively learn from challenging samples, where no correct answer exists within the sampled response, rendering the model unable to learn from such cases. 
In contrast, our difficulty-aware GRPO demonstrates superior generalization capabilities and is better equipped to learn from difficult samples by employing response resampling and advantage reweighting techniques.

\subsubsection{Effect of Two Training Stages}
% 消融first stage, the seconde stage
An ablation study evaluating two training stages is shown in Table \ref{tab:ab}. Skipping any training stage results in a decrease in performance, demonstrating the effectiveness of our two-stage training procedure.
On average, skipping Stage 1 results in a 0.89\% drop in accuracy, while omitting Stage 2 causes a significantly larger drop of 6.62\%.
This demonstrates that failing to align the soft prompts and contrastive embeddings during Stage 1, prior to fine-tuning the entire model, can degrade the performance of the MLLM. Moreover, fine-tuning with our difficulty-aware GRPO improves the model's suitability for the IAD task.

\subsubsection{Dataset}
% 数据集多出来 Object Descriptive Text as Query Image  效果。
We assess whether using object descriptive text as a query image in the training data can enhance performance. The results, presented in Table \ref{tab:ab}, show that incorporating text samples leads to a 2.58\% improvement in model accuracy. 
This improvement highlights the inherent alignment between text and images in MLLMs and demonstrates that using object descriptive text as a substitute for the query image effectively enriches the training data, thereby enhancing model performance.

\begin{table}[t]
	\caption{Ablation study on  Soft Prompt (SP)  and  Contrastive Embeddings (CE).}
      \vskip -0.1in 
 \label{tab:abs}
 \footnotesize
    \centering
    % \resizebox{\linewidth}{!}{
      \setlength{\tabcolsep}{1mm}{
   \begin{tabular}{ccc}
   \toprule
       &  Anomaly Discrimination  &  Defect Localization \\ \midrule
       w/o SP \& CE             & 73.51                                      & 75.22                                   \\
w/o SP                   & 73.63                                      & 76.07                                   \\
EMIT & \textbf{73.87}                                      & \textbf{76.39}            
 \\ 
 \bottomrule             
    \end{tabular}
    }
    % }
    \vskip -0.1in 
\end{table}

\subsubsection{Effect of the Soft Prompt and Contrastive Embeddings}
% 消融soft prompt + contrastive embedding
Table \ref{tab:abs} evaluates the impact of incorporating the soft prompt and contrastive embeddings on our model's performance. The results clearly demonstrate that removing either the soft prompt or contrastive embeddings results in a drop in performance, highlighting their critical contributions.
Specifically, omitting the soft prompt (w/o SP) leads to a more rigid connection between the task-question input and heatmap-guided contrastive embeddings, which negatively affects model performance. Furthermore, when both the soft prompt and contrastive embeddings are removed (w/o SP \& CE), the performance suffers an even greater decline. This emphasizes the importance of contrastive embeddings in providing valuable information that aids the model in detecting and localizing anomalies effectively.

\section{Conclusion}
In this paper, we present EMIT, a novel approach for enhancing MLLMs in  IAD through difficulty-aware GRPO.
Our method incorporates a soft prompt and contrastive embeddings into the MLLM to improve its performance in few-shot anomaly detection scenarios. 
By constructing a multi-task IAD dataset from seven sub-datasets and fine-tuning InterVL3-8B, we achieve an average improvement of 7.77\% over the base model across the seven tasks in MMAD.


\begin{thebibliography}{54}
\providecommand{\natexlab}[1]{#1}

\bibitem[{Achiam et~al.(2023)Achiam, Adler, Agarwal, Ahmad, Akkaya, Aleman,
  Almeida, Altenschmidt, Altman, Anadkat et~al.}]{achiam2023gpt}
Achiam, J.; Adler, S.; Agarwal, S.; Ahmad, L.; Akkaya, I.; Aleman, F.~L.;
  Almeida, D.; Altenschmidt, J.; Altman, S.; Anadkat, S.; et~al. 2023.
\newblock Gpt-4 technical report.
\newblock \emph{arXiv preprint arXiv:2303.08774}.

\bibitem[{Agrawal et~al.(2024)Agrawal, Antoniak, Hanna, Bout, Chaplot,
  Chudnovsky, Costa, Monicault, Garg, Gervet, Ghosh, Héliou, Jacob, Jiang,
  Khandelwal, Lacroix, Lample, Casas, Lavril, Scao, Lo, Marshall, Martin,
  Mensch, Muddireddy, Nemychnikova, Pellat, Platen, Raghuraman, Rozière,
  Sablayrolles, Saulnier, Sauvestre, Shang, Soletskyi, Stewart, Stock, Studnia,
  Subramanian, Vaze, Wang, and Yang}]{agrawal2024pixtral12b}
Agrawal, P.; Antoniak, S.; Hanna, E.~B.; Bout, B.; Chaplot, D.; Chudnovsky, J.;
  Costa, D.; Monicault, B.~D.; Garg, S.; Gervet, T.; Ghosh, S.; Héliou, A.;
  Jacob, P.; Jiang, A.~Q.; Khandelwal, K.; Lacroix, T.; Lample, G.; Casas,
  D.~L.; Lavril, T.; Scao, T.~L.; Lo, A.; Marshall, W.; Martin, L.; Mensch, A.;
  Muddireddy, P.; Nemychnikova, V.; Pellat, M.; Platen, P.~V.; Raghuraman, N.;
  Rozière, B.; Sablayrolles, A.; Saulnier, L.; Sauvestre, R.; Shang, W.;
  Soletskyi, R.; Stewart, L.; Stock, P.; Studnia, J.; Subramanian, S.; Vaze,
  S.; Wang, T.; and Yang, S. 2024.
\newblock Pixtral 12B.
\newblock arXiv:2410.07073.

\bibitem[{Bai et~al.(2023)Bai, Mou, Likhomanenko, Cinbis, Tuzel, Huang, Shan,
  Shi, and Cao}]{bai2023vision}
Bai, H.; Mou, S.; Likhomanenko, T.; Cinbis, R.~G.; Tuzel, O.; Huang, P.; Shan,
  J.; Shi, J.; and Cao, M. 2023.
\newblock Vision datasets: A benchmark for vision-based industrial inspection.
\newblock \emph{arXiv preprint arXiv:2306.07890}.

\bibitem[{Bai et~al.(2025)Bai, Chen, Liu, Wang, Ge, Song, Dang, Wang, Wang,
  Tang, Zhong, Zhu, Yang, Li, Wan, Wang, Ding, Fu, Xu, Ye, Zhang, Xie, Cheng,
  Zhang, Yang, Xu, and Lin}]{bai2025qwen25vltechnicalreport}
Bai, S.; Chen, K.; Liu, X.; Wang, J.; Ge, W.; Song, S.; Dang, K.; Wang, P.;
  Wang, S.; Tang, J.; Zhong, H.; Zhu, Y.; Yang, M.; Li, Z.; Wan, J.; Wang, P.;
  Ding, W.; Fu, Z.; Xu, Y.; Ye, J.; Zhang, X.; Xie, T.; Cheng, Z.; Zhang, H.;
  Yang, Z.; Xu, H.; and Lin, J. 2025.
\newblock Qwen2.5-VL Technical Report.
\newblock arXiv:2502.13923.

\bibitem[{Bergmann et~al.(2022)Bergmann, Batzner, Fauser, Sattlegger, and
  Steger}]{bergmann2022beyond}
Bergmann, P.; Batzner, K.; Fauser, M.; Sattlegger, D.; and Steger, C. 2022.
\newblock Beyond dents and scratches: Logical constraints in unsupervised
  anomaly detection and localization.
\newblock \emph{International Journal of Computer Vision}, 130(4): 947--969.

\bibitem[{Bergmann et~al.(2019)Bergmann, Fauser, Sattlegger, and
  Steger}]{bergmann2019mvtec}
Bergmann, P.; Fauser, M.; Sattlegger, D.; and Steger, C. 2019.
\newblock MVTec AD--A comprehensive real-world dataset for unsupervised anomaly
  detection.
\newblock In \emph{Proceedings of the IEEE/CVF conference on computer vision
  and pattern recognition}, 9592--9600.

\bibitem[{Cao et~al.(2024)Cao, Zhang, Frittoli, Cheng, Shen, and
  Boracchi}]{cao2024adaclip}
Cao, Y.; Zhang, J.; Frittoli, L.; Cheng, Y.; Shen, W.; and Boracchi, G. 2024.
\newblock Adaclip: Adapting clip with hybrid learnable prompts for zero-shot
  anomaly detection.
\newblock In \emph{European Conference on Computer Vision}, 55--72. Springer.

\bibitem[{Chao et~al.(2025)Chao, Liu, Tang, and Wu}]{chao2025anomalyr1}
Chao, Y.; Liu, J.; Tang, J.; and Wu, G. 2025.
\newblock Anomalyr1: A grpo-based end-to-end mllm for industrial anomaly
  detection.
\newblock \emph{arXiv preprint arXiv:2504.11914}.

\bibitem[{Chen et~al.(2025{\natexlab{a}})Chen, Tu, Wang, Liu, Tang, Du, Zhou,
  and Xie}]{chen2025sft}
Chen, H.; Tu, H.; Wang, F.; Liu, H.; Tang, X.; Du, X.; Zhou, Y.; and Xie, C.
  2025{\natexlab{a}}.
\newblock Sft or rl? an early investigation into training r1-like reasoning
  large vision-language models.
\newblock \emph{arXiv preprint arXiv:2504.11468}.

\bibitem[{Chen et~al.(2025{\natexlab{b}})Chen, Chen, Imani, and
  Imani}]{chen2025can}
Chen, Z.; Chen, H.; Imani, M.; and Imani, F. 2025{\natexlab{b}}.
\newblock Can multimodal large language models be guided to improve industrial
  anomaly detection?
\newblock \emph{arXiv preprint arXiv:2501.15795}.

\bibitem[{Deng et~al.(2024)Deng, Luo, Zhai, Cao, and Kang}]{deng2024vmad}
Deng, H.; Luo, H.; Zhai, W.; Cao, Y.; and Kang, Y. 2024.
\newblock Vmad: Visual-enhanced multimodal large language model for zero-shot
  anomaly detection.
\newblock \emph{arXiv preprint arXiv:2409.20146}.

\bibitem[{Dettmers et~al.(2024)Dettmers, Pagnoni, Holtzman, and
  Zettlemoyer}]{dettmers2024qlora}
Dettmers, T.; Pagnoni, A.; Holtzman, A.; and Zettlemoyer, L. 2024.
\newblock Qlora: Efficient finetuning of quantized llms.
\newblock \emph{Advances in Neural Information Processing Systems}, 36.

\bibitem[{Fan et~al.(2024)Fan, Huang, Di, Su, Pagnucco, and
  Song}]{fan2024revitalizing}
Fan, L.; Huang, J.; Di, D.; Su, A.; Pagnucco, M.; and Song, Y. 2024.
\newblock Revitalizing Reconstruction Models for Multi-class Anomaly Detection
  via Class-Aware Contrastive Learning.
\newblock \emph{arXiv preprint arXiv:2412.04769}.

\bibitem[{Fang et~al.(2023)Fang, Wang, Li, Liu, Hu, and
  Xiao}]{fang2023fastrecon}
Fang, Z.; Wang, X.; Li, H.; Liu, J.; Hu, Q.; and Xiao, J. 2023.
\newblock Fastrecon: Few-shot industrial anomaly detection via fast feature
  reconstruction.
\newblock In \emph{Proceedings of the IEEE/CVF International Conference on
  Computer Vision}, 17481--17490.

\bibitem[{Gu et~al.(2024)Gu, Zhu, Zhu, Chen, Tang, and Wang}]{gu2024anomalygpt}
Gu, Z.; Zhu, B.; Zhu, G.; Chen, Y.; Tang, M.; and Wang, J. 2024.
\newblock Anomalygpt: Detecting industrial anomalies using large
  vision-language models.
\newblock In \emph{Proceedings of the AAAI Conference on Artificial
  Intelligence}, volume~38, 1932--1940.

\bibitem[{Guo et~al.(2025)Guo, Lu, Zhang, Chen, Li, and Liao}]{guo2025dinomaly}
Guo, J.; Lu, S.; Zhang, W.; Chen, F.; Li, H.; and Liao, H. 2025.
\newblock Dinomaly: The less is more philosophy in multi-class unsupervised
  anomaly detection.
\newblock In \emph{Proceedings of the Computer Vision and Pattern Recognition
  Conference}, 20405--20415.

\bibitem[{Hoang et~al.(2025)Hoang, Tan, Nguyen, Tran, Duong, Mai, Pham, Phan,
  Do, Duong et~al.}]{hoang2025unsupervised}
Hoang, D.-C.; Tan, P.~X.; Nguyen, A.-N.; Tran, D.-T.; Duong, V.-H.; Mai, A.-T.;
  Pham, D.-L.; Phan, K.-T.; Do, M.-Q.; Duong, T. H.~A.; et~al. 2025.
\newblock Unsupervised visual-to-geometric feature reconstruction for
  vision-based industrial anomaly detection.
\newblock \emph{IEEE Access}.

\bibitem[{Hu et~al.(2025)Hu, Wang, Fan, Zeng, Lu, Hong, and
  Zhang}]{hu2025dsmbad}
Hu, H.; Wang, X.; Fan, J.; Zeng, Z.; Lu, J.; Hong, O.; and Zhang, J. 2025.
\newblock DSMBAD: Dual-Stream Memory Bank Framework for Unified Industrial
  Anomaly Detection.
\newblock \emph{Electronics}, 14(14): 2748.

\bibitem[{Hyun et~al.(2024)Hyun, Kim, Jeon, Kim, Bae, and
  Kang}]{hyun2024reconpatch}
Hyun, J.; Kim, S.; Jeon, G.; Kim, S.~H.; Bae, K.; and Kang, B.~J. 2024.
\newblock Reconpatch: Contrastive patch representation learning for industrial
  anomaly detection.
\newblock In \emph{Proceedings of the IEEE/CVF Winter Conference on
  Applications of Computer Vision}, 2052--2061.

\bibitem[{Jeong et~al.(2023)Jeong, Zou, Kim, Zhang, Ravichandran, and
  Dabeer}]{jeong2023winclip}
Jeong, J.; Zou, Y.; Kim, T.; Zhang, D.; Ravichandran, A.; and Dabeer, O. 2023.
\newblock Winclip: Zero-/few-shot anomaly classification and segmentation.
\newblock In \emph{Proceedings of the IEEE/CVF Conference on Computer Vision
  and Pattern Recognition}, 19606--19616.

\bibitem[{Jezek et~al.(2021)Jezek, Jonak, Burget, Dvorak, and
  Skotak}]{jezek2021deep}
Jezek, S.; Jonak, M.; Burget, R.; Dvorak, P.; and Skotak, M. 2021.
\newblock Deep learning-based defect detection of metal parts: evaluating
  current methods in complex conditions.
\newblock In \emph{2021 13th International congress on ultra modern
  telecommunications and control systems and workshops (ICUMT)}, 66--71. IEEE.

\bibitem[{Jiang et~al.(2024{\natexlab{a}})Jiang, Li, Deng, Liu, Gao, Zhou, Li,
  Wang, and Zheng}]{jiang2024mmad}
Jiang, X.; Li, J.; Deng, H.; Liu, Y.; Gao, B.-B.; Zhou, Y.; Li, J.; Wang, C.;
  and Zheng, F. 2024{\natexlab{a}}.
\newblock Mmad: A comprehensive benchmark for multimodal large language models
  in industrial anomaly detection.
\newblock \emph{arXiv preprint arXiv:2410.09453}.

\bibitem[{Jiang et~al.(2022)Jiang, Liu, Wang, Nie, Wu, Liu, Wang, and
  Zheng}]{jiang2022softpatch}
Jiang, X.; Liu, J.; Wang, J.; Nie, Q.; Wu, K.; Liu, Y.; Wang, C.; and Zheng, F.
  2022.
\newblock Softpatch: Unsupervised anomaly detection with noisy data.
\newblock \emph{Advances in Neural Information Processing Systems}, 35:
  15433--15445.

\bibitem[{Jiang et~al.(2024{\natexlab{b}})Jiang, Lu, Jin, Sun, Wu, and
  Zhuo}]{jiang2024fabgpt}
Jiang, Y.; Lu, X.; Jin, Q.; Sun, Q.; Wu, H.; and Zhuo, C. 2024{\natexlab{b}}.
\newblock Fabgpt: An efficient large multimodal model for complex wafer defect
  knowledge queries.
\newblock In \emph{Proceedings of the 43rd IEEE/ACM International Conference on
  Computer-Aided Design}, 1--8.

\bibitem[{Jin et~al.(2025)Jin, Feng, Mou, Lakemeyer, Decker, Simons, and
  Stegmaier}]{jin2025logicad}
Jin, E.; Feng, Q.; Mou, Y.; Lakemeyer, G.; Decker, S.; Simons, O.; and
  Stegmaier, J. 2025.
\newblock Logicad: Explainable anomaly detection via vlm-based text feature
  extraction.
\newblock In \emph{Proceedings of the AAAI Conference on Artificial
  Intelligence}, volume~39, 4129--4137.

\bibitem[{Kim et~al.(2025)Kim, Park, Cho, Lim, Kang, Lee, and
  Lee}]{kim2025genclip}
Kim, D.; Park, C.; Cho, S.; Lim, H.; Kang, M.; Lee, J.; and Lee, S. 2025.
\newblock GenCLIP: Generalizing CLIP Prompts for Zero-shot Anomaly Detection.
\newblock \emph{arXiv preprint arXiv:2504.14919}.

\bibitem[{Li et~al.(2024{\natexlab{a}})Li, Zhang, Guo, Zhang, Li, Zhang, Zhang,
  Zhang, Li, Liu, and Li}]{li2024llavaonevisioneasyvisualtask}
Li, B.; Zhang, Y.; Guo, D.; Zhang, R.; Li, F.; Zhang, H.; Zhang, K.; Zhang, P.;
  Li, Y.; Liu, Z.; and Li, C. 2024{\natexlab{a}}.
\newblock LLaVA-OneVision: Easy Visual Task Transfer.
\newblock arXiv:2408.03326.

\bibitem[{Li et~al.(2024{\natexlab{b}})Li, Zhang, Zhang, Zhang, Li, Li, Ma, and
  Li}]{li2024llavanextinterleavetacklingmultiimagevideo}
Li, F.; Zhang, R.; Zhang, H.; Zhang, Y.; Li, B.; Li, W.; Ma, Z.; and Li, C.
  2024{\natexlab{b}}.
\newblock LLaVA-NeXT-Interleave: Tackling Multi-image, Video, and 3D in Large
  Multimodal Models.
\newblock arXiv:2407.07895.

\bibitem[{Li et~al.(2024{\natexlab{c}})Li, He, Ying, Li, and
  Zhou}]{li2024memadet}
Li, M.; He, J.; Ying, Z.; Li, G.; and Zhou, M. 2024{\natexlab{c}}.
\newblock MemADet: A Representative Memory Bank Approach for Industrial Image
  Anomaly Detection.
\newblock In \emph{2024 IEEE International Conference on Systems, Man, and
  Cybernetics (SMC)}, 2261--2266. IEEE.

\bibitem[{Li et~al.(2025)Li, Chu, Chen, Xie, Shan, and Zhao}]{li2025lad}
Li, W.; Chu, G.; Chen, J.; Xie, G.-S.; Shan, C.; and Zhao, F. 2025.
\newblock Lad-reasoner: Tiny multimodal models are good reasoners for logical
  anomaly detection.
\newblock \emph{arXiv preprint arXiv:2504.12749}.

\bibitem[{Li et~al.(2023)Li, Wang, Yuan, Liu, Zhao, Guo, Xu, Shi, and
  Zuo}]{li2023myriad}
Li, Y.; Wang, H.; Yuan, S.; Liu, M.; Zhao, D.; Guo, Y.; Xu, C.; Shi, G.; and
  Zuo, W. 2023.
\newblock Myriad: Large multimodal model by applying vision experts for
  industrial anomaly detection.
\newblock \emph{arXiv preprint arXiv:2310.19070}.

\bibitem[{Loshchilov(2017)}]{loshchilov2017decoupled}
Loshchilov, I. 2017.
\newblock Decoupled weight decay regularization.
\newblock \emph{arXiv preprint arXiv:1711.05101}.

\bibitem[{Mokhtar et~al.(2025)Mokhtar, Mousakhan, Galesso, Tayyub, and
  Brox}]{mokhtar2025detect}
Mokhtar, S.; Mousakhan, A.; Galesso, S.; Tayyub, J.; and Brox, T. 2025.
\newblock Detect, Classify, Act: Categorizing Industrial Anomalies with
  Multi-Modal Large Language Models.
\newblock In \emph{Proceedings of the Computer Vision and Pattern Recognition
  Conference}, 4058--4067.

\bibitem[{Roth et~al.(2022)Roth, Pemula, Zepeda, Sch{\"o}lkopf, Brox, and
  Gehler}]{roth2022towards}
Roth, K.; Pemula, L.; Zepeda, J.; Sch{\"o}lkopf, B.; Brox, T.; and Gehler, P.
  2022.
\newblock Towards total recall in industrial anomaly detection.
\newblock In \emph{Proceedings of the IEEE/CVF conference on computer vision
  and pattern recognition}, 14318--14328.

\bibitem[{Schulman et~al.(2017)Schulman, Wolski, Dhariwal, Radford, and
  Klimov}]{schulman2017proximal}
Schulman, J.; Wolski, F.; Dhariwal, P.; Radford, A.; and Klimov, O. 2017.
\newblock Proximal policy optimization algorithms.
\newblock \emph{arXiv preprint arXiv:1707.06347}.

\bibitem[{Shao et~al.(2024)Shao, Wang, Zhu, Xu, Song, Bi, Zhang, Zhang, Li, Wu
  et~al.}]{shao2024deepseekmath}
Shao, Z.; Wang, P.; Zhu, Q.; Xu, R.; Song, J.; Bi, X.; Zhang, H.; Zhang, M.;
  Li, Y.; Wu, Y.; et~al. 2024.
\newblock Deepseekmath: Pushing the limits of mathematical reasoning in open
  language models.
\newblock \emph{arXiv preprint arXiv:2402.03300}.

\bibitem[{Team et~al.(2025{\natexlab{a}})Team, Kamath, Ferret, Pathak,
  Vieillard, Merhej, Perrin, Matejovicova, Ramé, Rivière, Rouillard, Mesnard,
  Cideron, bastien Grill, Ramos, Yvinec, Casbon, Pot, Penchev, Liu, Visin,
  Kenealy, Beyer, Zhai, Tsitsulin, Busa-Fekete, Feng, Sachdeva, Coleman, Gao,
  Mustafa, Barr, Parisotto, Tian, Eyal, Cherry, Peter, Sinopalnikov,
  Bhupatiraju, Agarwal, Kazemi, Malkin, Kumar, Vilar, Brusilovsky, Luo,
  Steiner, Friesen, Sharma, Sharma, Gilady, Goedeckemeyer, Saade, Feng,
  Kolesnikov, Bendebury, Abdagic, Vadi, György, Pinto, Das, Bapna, Miech,
  Yang, Paterson, Shenoy, Chakrabarti, Piot, Wu, Shahriari, Petrini, Chen, Lan,
  Choquette-Choo, Carey, Brick, Deutsch, Eisenbud, Cattle, Cheng, Paparas,
  Sreepathihalli, Reid, Tran, Zelle, Noland, Huizenga, Kharitonov, Liu,
  Amirkhanyan, Cameron, Hashemi, Klimczak-Plucińska, Singh, Mehta, Lehri,
  Hazimeh, Ballantyne, Szpektor, Nardini, Pouget-Abadie, Chan, Stanton,
  Wieting, Lai, Orbay, Fernandez, Newlan, yeong Ji, Singh, Black, Yu, Hui,
  Vodrahalli, Greff, Qiu, Valentine, Coelho, Ritter, Hoffman, Watson,
  Chaturvedi, Moynihan, Ma, Babar, Noy, Byrd, Roy, Momchev, Chauhan, Sachdeva,
  Bunyan, Botarda, Caron, Rubenstein, Culliton, Schmid, Sessa, Xu, Stanczyk,
  Tafti, Shivanna, Wu, Pan, Rokni, Willoughby, Vallu, Mullins, Jerome, Smoot,
  Girgin, Iqbal, Reddy, Sheth, Põder, Bhatnagar, Panyam, Eiger, Zhang, Liu,
  Yacovone, Liechty, Kalra, Evci, Misra, Roseberry, Feinberg, Kolesnikov, Han,
  Kwon, Chen, Chow, Zhu, Wei, Egyed, Cotruta, Giang, Kirk, Rao, Black, Babar,
  Lo, Moreira, Martins, Sanseviero, Gonzalez, Gleicher, Warkentin, Mirrokni,
  Senter, Collins, Barral, Ghahramani, Hadsell, Matias, Sculley, Petrov,
  Fiedel, Shazeer, Vinyals, Dean, Hassabis, Kavukcuoglu, Farabet, Buchatskaya,
  Alayrac, Anil, Dmitry, Lepikhin, Borgeaud, Bachem, Joulin, Andreev, Hardin,
  Dadashi, and Hussenot}]{gemmateam2025gemma3technicalreport}
Team, G.; Kamath, A.; Ferret, J.; Pathak, S.; Vieillard, N.; Merhej, R.;
  Perrin, S.; Matejovicova, T.; Ramé, A.; Rivière, M.; Rouillard, L.;
  Mesnard, T.; Cideron, G.; bastien Grill, J.; Ramos, S.; Yvinec, E.; Casbon,
  M.; Pot, E.; Penchev, I.; Liu, G.; Visin, F.; Kenealy, K.; Beyer, L.; Zhai,
  X.; Tsitsulin, A.; Busa-Fekete, R.; Feng, A.; Sachdeva, N.; Coleman, B.; Gao,
  Y.; Mustafa, B.; Barr, I.; Parisotto, E.; Tian, D.; Eyal, M.; Cherry, C.;
  Peter, J.-T.; Sinopalnikov, D.; Bhupatiraju, S.; Agarwal, R.; Kazemi, M.;
  Malkin, D.; Kumar, R.; Vilar, D.; Brusilovsky, I.; Luo, J.; Steiner, A.;
  Friesen, A.; Sharma, A.; Sharma, A.; Gilady, A.~M.; Goedeckemeyer, A.; Saade,
  A.; Feng, A.; Kolesnikov, A.; Bendebury, A.; Abdagic, A.; Vadi, A.; György,
  A.; Pinto, A.~S.; Das, A.; Bapna, A.; Miech, A.; Yang, A.; Paterson, A.;
  Shenoy, A.; Chakrabarti, A.; Piot, B.; Wu, B.; Shahriari, B.; Petrini, B.;
  Chen, C.; Lan, C.~L.; Choquette-Choo, C.~A.; Carey, C.; Brick, C.; Deutsch,
  D.; Eisenbud, D.; Cattle, D.; Cheng, D.; Paparas, D.; Sreepathihalli, D.~S.;
  Reid, D.; Tran, D.; Zelle, D.; Noland, E.; Huizenga, E.; Kharitonov, E.; Liu,
  F.; Amirkhanyan, G.; Cameron, G.; Hashemi, H.; Klimczak-Plucińska, H.;
  Singh, H.; Mehta, H.; Lehri, H.~T.; Hazimeh, H.; Ballantyne, I.; Szpektor,
  I.; Nardini, I.; Pouget-Abadie, J.; Chan, J.; Stanton, J.; Wieting, J.; Lai,
  J.; Orbay, J.; Fernandez, J.; Newlan, J.; yeong Ji, J.; Singh, J.; Black, K.;
  Yu, K.; Hui, K.; Vodrahalli, K.; Greff, K.; Qiu, L.; Valentine, M.; Coelho,
  M.; Ritter, M.; Hoffman, M.; Watson, M.; Chaturvedi, M.; Moynihan, M.; Ma,
  M.; Babar, N.; Noy, N.; Byrd, N.; Roy, N.; Momchev, N.; Chauhan, N.;
  Sachdeva, N.; Bunyan, O.; Botarda, P.; Caron, P.; Rubenstein, P.~K.;
  Culliton, P.; Schmid, P.; Sessa, P.~G.; Xu, P.; Stanczyk, P.; Tafti, P.;
  Shivanna, R.; Wu, R.; Pan, R.; Rokni, R.; Willoughby, R.; Vallu, R.; Mullins,
  R.; Jerome, S.; Smoot, S.; Girgin, S.; Iqbal, S.; Reddy, S.; Sheth, S.;
  Põder, S.; Bhatnagar, S.; Panyam, S.~R.; Eiger, S.; Zhang, S.; Liu, T.;
  Yacovone, T.; Liechty, T.; Kalra, U.; Evci, U.; Misra, V.; Roseberry, V.;
  Feinberg, V.; Kolesnikov, V.; Han, W.; Kwon, W.; Chen, X.; Chow, Y.; Zhu, Y.;
  Wei, Z.; Egyed, Z.; Cotruta, V.; Giang, M.; Kirk, P.; Rao, A.; Black, K.;
  Babar, N.; Lo, J.; Moreira, E.; Martins, L.~G.; Sanseviero, O.; Gonzalez, L.;
  Gleicher, Z.; Warkentin, T.; Mirrokni, V.; Senter, E.; Collins, E.; Barral,
  J.; Ghahramani, Z.; Hadsell, R.; Matias, Y.; Sculley, D.; Petrov, S.; Fiedel,
  N.; Shazeer, N.; Vinyals, O.; Dean, J.; Hassabis, D.; Kavukcuoglu, K.;
  Farabet, C.; Buchatskaya, E.; Alayrac, J.-B.; Anil, R.; Dmitry; Lepikhin;
  Borgeaud, S.; Bachem, O.; Joulin, A.; Andreev, A.; Hardin, C.; Dadashi, R.;
  and Hussenot, L. 2025{\natexlab{a}}.
\newblock Gemma 3 Technical Report.
\newblock arXiv:2503.19786.

\bibitem[{Team et~al.(2025{\natexlab{b}})Team, Du, Yin, Xing, Qu, Wang, Chen,
  Zhang, Du, Wei, Wang, Zhang, Du, Wang, Yuan, Lu, Li, Sung, Wei, Lai, Zhu,
  Ding, Hu, Yang, Zhang, Wu, Yao, Lu, Wang, Gao, Zheng, Li, Su, Wang, Deng,
  Qiu, Xie, Wang, Liu, Yan, Ouyang, Chen, Sui, Yu, Dong, Dong, Xu, Cheng, Gu,
  Zhou, Liu, Cao, Yu, Song, Bai, Song, He, Huang, Xu, Yuan, Yao, Wu, Li, Zu,
  Zhou, Wang, Charles, Zhong, Li, Hu, Chen, Wang, Liu, Miao, Qin, Chen, Bao,
  Wang, Kang, Liu, Dong, Du, Wu, Wang, Yan, Zhou, Li, Jiang, Zhang, Yang,
  Huang, Huang, Zhao, Chen, and Lin}]{kimiteam2025kimivltechnicalreport}
Team, K.; Du, A.; Yin, B.; Xing, B.; Qu, B.; Wang, B.; Chen, C.; Zhang, C.; Du,
  C.; Wei, C.; Wang, C.; Zhang, D.; Du, D.; Wang, D.; Yuan, E.; Lu, E.; Li, F.;
  Sung, F.; Wei, G.; Lai, G.; Zhu, H.; Ding, H.; Hu, H.; Yang, H.; Zhang, H.;
  Wu, H.; Yao, H.; Lu, H.; Wang, H.; Gao, H.; Zheng, H.; Li, J.; Su, J.; Wang,
  J.; Deng, J.; Qiu, J.; Xie, J.; Wang, J.; Liu, J.; Yan, J.; Ouyang, K.; Chen,
  L.; Sui, L.; Yu, L.; Dong, M.; Dong, M.; Xu, N.; Cheng, P.; Gu, Q.; Zhou, R.;
  Liu, S.; Cao, S.; Yu, T.; Song, T.; Bai, T.; Song, W.; He, W.; Huang, W.; Xu,
  W.; Yuan, X.; Yao, X.; Wu, X.; Li, X.; Zu, X.; Zhou, X.; Wang, X.; Charles,
  Y.; Zhong, Y.; Li, Y.; Hu, Y.; Chen, Y.; Wang, Y.; Liu, Y.; Miao, Y.; Qin,
  Y.; Chen, Y.; Bao, Y.; Wang, Y.; Kang, Y.; Liu, Y.; Dong, Y.; Du, Y.; Wu, Y.;
  Wang, Y.; Yan, Y.; Zhou, Z.; Li, Z.; Jiang, Z.; Zhang, Z.; Yang, Z.; Huang,
  Z.; Huang, Z.; Zhao, Z.; Chen, Z.; and Lin, Z. 2025{\natexlab{b}}.
\newblock Kimi-VL Technical Report.
\newblock arXiv:2504.07491.

\bibitem[{Team et~al.(2025{\natexlab{c}})Team, Hong, Yu, Gu, Wang, Gan, Tang,
  Cheng, Qi, Ji, Pan, Duan, Wang, Wang, Cheng, He, Su, Yang, Pan, Zeng, Wang,
  Shi, Pang, Zhang, Yin, Yang, Chen, Xu, Chen, Chen, Chen, Lin, Wang, Chen,
  Lei, Gong, Pan, Zhang, Zheng, Yang, Zhong, Huang, Zhao, Xue, Tu, Meng, Zhang,
  Luo, Hao, Li, Jia, Lyu, Huang, Wang, Xue, Wang, An, Du, Shi, Huang, Niu,
  Wang, Yue, Li, Zhang, Zhang, Du, Hou, Xue, Du, Wang, Zhang, Liu, Xu, Li,
  Huang, Dong, and Tang}]{vteam2025glm41vthinkingversatilemultimodalreasoning}
Team, V.; Hong, W.; Yu, W.; Gu, X.; Wang, G.; Gan, G.; Tang, H.; Cheng, J.; Qi,
  J.; Ji, J.; Pan, L.; Duan, S.; Wang, W.; Wang, Y.; Cheng, Y.; He, Z.; Su, Z.;
  Yang, Z.; Pan, Z.; Zeng, A.; Wang, B.; Shi, B.; Pang, C.; Zhang, C.; Yin, D.;
  Yang, F.; Chen, G.; Xu, J.; Chen, J.; Chen, J.; Chen, J.; Lin, J.; Wang, J.;
  Chen, J.; Lei, L.; Gong, L.; Pan, L.; Zhang, M.; Zheng, Q.; Yang, S.; Zhong,
  S.; Huang, S.; Zhao, S.; Xue, S.; Tu, S.; Meng, S.; Zhang, T.; Luo, T.; Hao,
  T.; Li, W.; Jia, W.; Lyu, X.; Huang, X.; Wang, Y.; Xue, Y.; Wang, Y.; An, Y.;
  Du, Y.; Shi, Y.; Huang, Y.; Niu, Y.; Wang, Y.; Yue, Y.; Li, Y.; Zhang, Y.;
  Zhang, Y.; Du, Z.; Hou, Z.; Xue, Z.; Du, Z.; Wang, Z.; Zhang, P.; Liu, D.;
  Xu, B.; Li, J.; Huang, M.; Dong, Y.; and Tang, J. 2025{\natexlab{c}}.
\newblock GLM-4.1V-Thinking: Towards Versatile Multimodal Reasoning with
  Scalable Reinforcement Learning.
\newblock arXiv:2507.01006.

\bibitem[{Wang et~al.(2024)Wang, Zhu, Gao, Gan, Zhang, Gu, Qian, Chen, and
  Ma}]{wang2024real}
Wang, C.; Zhu, W.; Gao, B.-B.; Gan, Z.; Zhang, J.; Gu, Z.; Qian, S.; Chen, M.;
  and Ma, L. 2024.
\newblock Real-iad: A real-world multi-view dataset for benchmarking versatile
  industrial anomaly detection.
\newblock In \emph{Proceedings of the IEEE/CVF Conference on Computer Vision
  and Pattern Recognition}, 22883--22892.

\bibitem[{Xiaomi(2025)}]{coreteam2025mimovltechnicalreport}
Xiaomi, L.-C.-T. 2025.
\newblock MiMo-VL Technical Report.
\newblock arXiv:2506.03569.

\bibitem[{Xu et~al.(2025)Xu, Lo, Safaei, Patel, and Dwivedi}]{xu2025towards}
Xu, J.; Lo, S.-Y.; Safaei, B.; Patel, V.~M.; and Dwivedi, I. 2025.
\newblock Towards zero-shot anomaly detection and reasoning with multimodal
  large language models.
\newblock In \emph{Proceedings of the Computer Vision and Pattern Recognition
  Conference}, 20370--20382.

\bibitem[{Yeo et~al.(2025)Yeo, Tong, Niu, Neubig, and
  Yue}]{yeo2025demystifyinglongchainofthoughtreasoning}
Yeo, E.; Tong, Y.; Niu, M.; Neubig, G.; and Yue, X. 2025.
\newblock Demystifying Long Chain-of-Thought Reasoning in LLMs.
\newblock arXiv:2502.03373.

\bibitem[{Yu et~al.(2025)Yu, Zhang, Zhu, Yuan, Zuo, Yue, Dai, Fan, Liu, Liu
  et~al.}]{yu2025dapo}
Yu, Q.; Zhang, Z.; Zhu, R.; Yuan, Y.; Zuo, X.; Yue, Y.; Dai, W.; Fan, T.; Liu,
  G.; Liu, L.; et~al. 2025.
\newblock Dapo: An open-source llm reinforcement learning system at scale.
\newblock \emph{arXiv preprint arXiv:2503.14476}.

\bibitem[{Yuan et~al.(2025)Yuan, Jie, Zhang, Li, and Gao}]{yuan2025mfp}
Yuan, J.; Jie, P.; Zhang, J.; Li, Z.; and Gao, C. 2025.
\newblock MFP-CLIP: Exploring the Efficacy of Multi-Form Prompts for Zero-Shot
  Industrial Anomaly Detection.
\newblock \emph{arXiv preprint arXiv:2503.12910}.

\bibitem[{Zeng et~al.(2025)Zeng, Pang, Wang, and Yang}]{zeng2025lr}
Zeng, P.; Pang, F.; Wang, Z.; and Yang, A. 2025.
\newblock LR-IAD: Mask-Free Industrial Anomaly Detection with Logical
  Reasoning.
\newblock \emph{arXiv preprint arXiv:2504.19524}.

\bibitem[{Zhang et~al.(2024{\natexlab{a}})Zhang, Ding, Ban, and
  Dai}]{zhang2024pku}
Zhang, J.; Ding, R.; Ban, M.; and Dai, L. 2024{\natexlab{a}}.
\newblock PKU-GoodsAD: A supermarket goods dataset for unsupervised anomaly
  detection and segmentation.
\newblock \emph{IEEE Robotics and Automation Letters}, 9(3): 2008--2015.

\bibitem[{Zhang, Xu, and Zhou(2024)}]{zhang2024realnet}
Zhang, X.; Xu, M.; and Zhou, X. 2024.
\newblock Realnet: A feature selection network with realistic synthetic anomaly
  for anomaly detection.
\newblock In \emph{Proceedings of the IEEE/CVF conference on computer vision
  and pattern recognition}, 16699--16708.

\bibitem[{Zhang et~al.(2024{\natexlab{b}})Zhang, Cao, Xu, and
  Shen}]{zhang2024logicode}
Zhang, Y.; Cao, Y.; Xu, X.; and Shen, W. 2024{\natexlab{b}}.
\newblock Logicode: an llm-driven framework for logical anomaly detection.
\newblock \emph{IEEE Transactions on Automation Science and Engineering}.

\bibitem[{Zhang et~al.(2025)Zhang, Ruan, Gao, Liu, and Fu}]{zhang2025eiad}
Zhang, Z.; Ruan, J.; Gao, X.; Liu, T.; and Fu, Y. 2025.
\newblock Eiad: Explainable industrial anomaly detection via multi-modal large
  language models.
\newblock \emph{arXiv preprint arXiv:2503.14162}.

\bibitem[{Zhao et~al.(2025)Zhao, Lin, Han, Zhao, and Wei}]{zhao2025omniad}
Zhao, S.; Lin, Y.; Han, L.; Zhao, Y.; and Wei, Y. 2025.
\newblock OmniAD: Detect and Understand Industrial Anomaly via Multimodal
  Reasoning.
\newblock \emph{arXiv preprint arXiv:2505.22039}.

\bibitem[{Zhou et~al.(2023)Zhou, Pang, Tian, He, and
  Chen}]{zhou2023anomalyclip}
Zhou, Q.; Pang, G.; Tian, Y.; He, S.; and Chen, J. 2023.
\newblock Anomalyclip: Object-agnostic prompt learning for zero-shot anomaly
  detection.
\newblock \emph{arXiv preprint arXiv:2310.18961}.

\bibitem[{Zhu et~al.(2025)Zhu, Wang, Chen, Liu, Ye, Gu, Tian, Duan, Su, Shao,
  Gao, Cui, Wang, Cao, Liu, Wei, Zhang, Wang, Xu, Li, Wang, Deng, Li, He,
  Jiang, Luo, Wang, He, Shi, Zhang, Shao, He, Xiong, Qu, Sun, Jiao, Lv, Wu,
  Zhang, Deng, Ge, Chen, Wang, Dou, Lu, Zhu, Lu, Lin, Qiao, Dai, and
  Wang}]{zhu2025internvl3exploringadvancedtraining}
Zhu, J.; Wang, W.; Chen, Z.; Liu, Z.; Ye, S.; Gu, L.; Tian, H.; Duan, Y.; Su,
  W.; Shao, J.; Gao, Z.; Cui, E.; Wang, X.; Cao, Y.; Liu, Y.; Wei, X.; Zhang,
  H.; Wang, H.; Xu, W.; Li, H.; Wang, J.; Deng, N.; Li, S.; He, Y.; Jiang, T.;
  Luo, J.; Wang, Y.; He, C.; Shi, B.; Zhang, X.; Shao, W.; He, J.; Xiong, Y.;
  Qu, W.; Sun, P.; Jiao, P.; Lv, H.; Wu, L.; Zhang, K.; Deng, H.; Ge, J.; Chen,
  K.; Wang, L.; Dou, M.; Lu, L.; Zhu, X.; Lu, T.; Lin, D.; Qiao, Y.; Dai, J.;
  and Wang, W. 2025.
\newblock InternVL3: Exploring Advanced Training and Test-Time Recipes for
  Open-Source Multimodal Models.
\newblock arXiv:2504.10479.

\bibitem[{Zou et~al.(2022)Zou, Jeong, Pemula, Zhang, and Dabeer}]{zou2022spot}
Zou, Y.; Jeong, J.; Pemula, L.; Zhang, D.; and Dabeer, O. 2022.
\newblock Spot-the-difference self-supervised pre-training for anomaly
  detection and segmentation.
\newblock In \emph{European conference on computer vision}, 392--408. Springer.

\end{thebibliography}
\end{document}